\documentclass[letterpaper]{article} %
\usepackage{aaai23}  %
\usepackage{times}  %
\usepackage{helvet}  %
\usepackage{courier}  %
\usepackage[hyphens]{url}  %
\usepackage{graphicx} %
\usepackage{natbib}  %
\usepackage{caption} %
\usepackage{algorithm}
\usepackage{algorithmic}
\usepackage{newfloat}
\usepackage{listings}
\DeclareCaptionStyle{ruled}{labelfont=normalfont,labelsep=colon,strut=off} %
\floatstyle{ruled}
\newfloat{listing}{tb}{lst}{}
\floatname{listing}{Listing}
\usepackage{url}            %
\usepackage{booktabs}       %
\usepackage{amsfonts}       %
\usepackage{nicefrac}       %
\usepackage{microtype}      %
\usepackage{xcolor}         %
\usepackage{amsmath}
\usepackage{amssymb}
\usepackage{mathtools}
\usepackage{amsthm}
\usepackage{epsfig}
\usepackage{graphicx}
\usepackage{url}
\usepackage{caption}%
\usepackage{subcaption}
\usepackage{multirow}
\usepackage{color}
\usepackage{xcolor}
\usepackage{misc_symbols}
\usepackage{enumitem}

\newcommand{\iri}{IRI}
\newcommand{\iris}{IRIs}

\title{Do Invariances in Deep Neural Networks Align with Human Perception?}

\author{%
  Vedant Nanda\thanks{Correspondence to: \texttt{vnanda@mpi-sws.org}}\textsuperscript{\rm 1 \rm 2},
  Ayan Majumdar\textsuperscript{\rm 2},
  Camila Kolling\textsuperscript{\rm 2},
  John P. Dickerson\textsuperscript{\rm 1},
  Krishna P. Gummadi\textsuperscript{\rm 2},
  Bradley C. Love\textsuperscript{\rm 3 \rm 4},
  Adrian Weller\textsuperscript{\rm 3 \rm 5},
}
\affiliations{
    \textsuperscript{\rm 1}University of Maryland \,\,\, \textsuperscript{\rm 2}MPI-SWS\\
    \textsuperscript{\rm 3}The Alan Turing Institute \,\,\, \textsuperscript{\rm 4}University College London \,\,\, \textsuperscript{\rm 5}University of Cambridge
}

\begin{document}

\maketitle

\begin{abstract}

An evaluation criterion for safe and trustworthy deep learning is how well the invariances captured by representations of deep neural networks (DNNs) are shared with humans. We identify challenges in measuring these invariances. Prior works used gradient-based methods to generate \textit{identically represented inputs} (\iris), \ie, inputs which have identical representations (on a given layer) of a neural network, and thus capture invariances of a given network. One necessary criterion for a network's invariances to align with human perception is for its \iris~look ``similar'' to humans. Prior works, however, have mixed takeaways; some argue that later layers of DNNs do not learn human-like invariances (\cite{jenelle2019metamers}) yet others seem to indicate otherwise (\cite{mahendran2014understanding}). We argue that the loss function used to generate \iris~can heavily affect takeaways about invariances of the network and is the primary reason for these conflicting findings. We propose an \textit{adversarial} regularizer on the \iri-generation loss that finds \iris~that make any model appear to have very little shared invariance with humans. Based on this evidence, we argue that there is scope for improving models to have human-like invariances, and further, to have meaningful comparisons between models one should use \iris~generated using the \textit{regularizer-free} loss.
We then conduct an in-depth investigation of how different components (\eg~architectures, training losses, data augmentations) of the deep learning pipeline contribute to learning models that have good alignment with humans. We find that architectures with residual connections trained using a (self-supervised) contrastive loss with $\ell_p$ ball adversarial data augmentation tend to learn invariances that are most aligned with humans. Code: 
\url{github.com/nvedant07/Human-NN-Alignment}

\end{abstract}

\section{Introduction}\label{sec:intro}

The ability to train deep neural networks (DNNs) which learn useful features and representations is key for their widespread use~\cite{bengio2013representation,lecun2015deep}. 
In domains where DNNs are used for tasks that previously required human intelligence (\eg~image classification) and where safety and trustworthiness are important considerations, it is helpful to assess the alignment of the learned representations with human perception.
Such assessments can help in understanding and diagnosing issues such as lack of robustness to distribution shifts~\cite{recht2019imagenet,taori2020measuring}, adversarial attacks~\cite{papernot2016limitations,goodfellow2015explaining} or using undesirable features for a downstream task~\cite{beery2018recognition,sagawa2019distributionally,rosenfeld2018elephant,buolamwini2018gender,ilyas2019adversarial}.

One test of human-machine alignment is whether different images that map to identical internal network representation are also judged as identical by humans. To study alignment with human perception, prior works have used the approach of \textit{representation inversion}~\cite{mahendran2014understanding}. 
The key idea is the following: given an input to a neural network, the approach first finds \textit{identically represented inputs} (IRIs), \ie~inputs which have similar representations on some given layer(s) of the neural network.
In the second step, the inputs that are perceived similarly by the neural network are checked by humans for visual similarity.
Thus, the approach relies on estimating whether a transformation of the inputs which is representation invariant to a neural network is also an invariant transformation to the human eye, \ie~it checks whether models and humans have shared or aligned invariances.

Prior works use gradient-based methods to generate IRIs for a given target input starting with a random seed input. 
These works revealed exciting insights: (a) Feather et al. studied representational invariance for different layers of DNNs trained over ImageNet data (using the standard cross-entropy loss). 
They showed that while later layer representations of  DNNs do not share any invariances with human perception, the earlier layers are somewhat better aligned with human perception~\cite{jenelle2019metamers}.
(b) Engstrom et al. found that, unlike standard DNNs, adversarially robust DNNs, \ie, DNNs trained using adversarial training~\cite{madry2019deep}, learn representations that are well aligned with human perception, even in later layers~\cite{engstrom2019adversarial}. This was also confirmed by other works~\cite{kaur2019perceptually,santurkar2019image}.
However, some of these findings are contradicted when differently regularized methods are used for generating IRIs, which show that even later layers of DNNs learn human aligned representations~\cite{mahendran2014understanding, olah2017feature, olah2020zoom}.

\begin{figure*}[t!]
    \centering
        \includegraphics[width=0.7\textwidth]{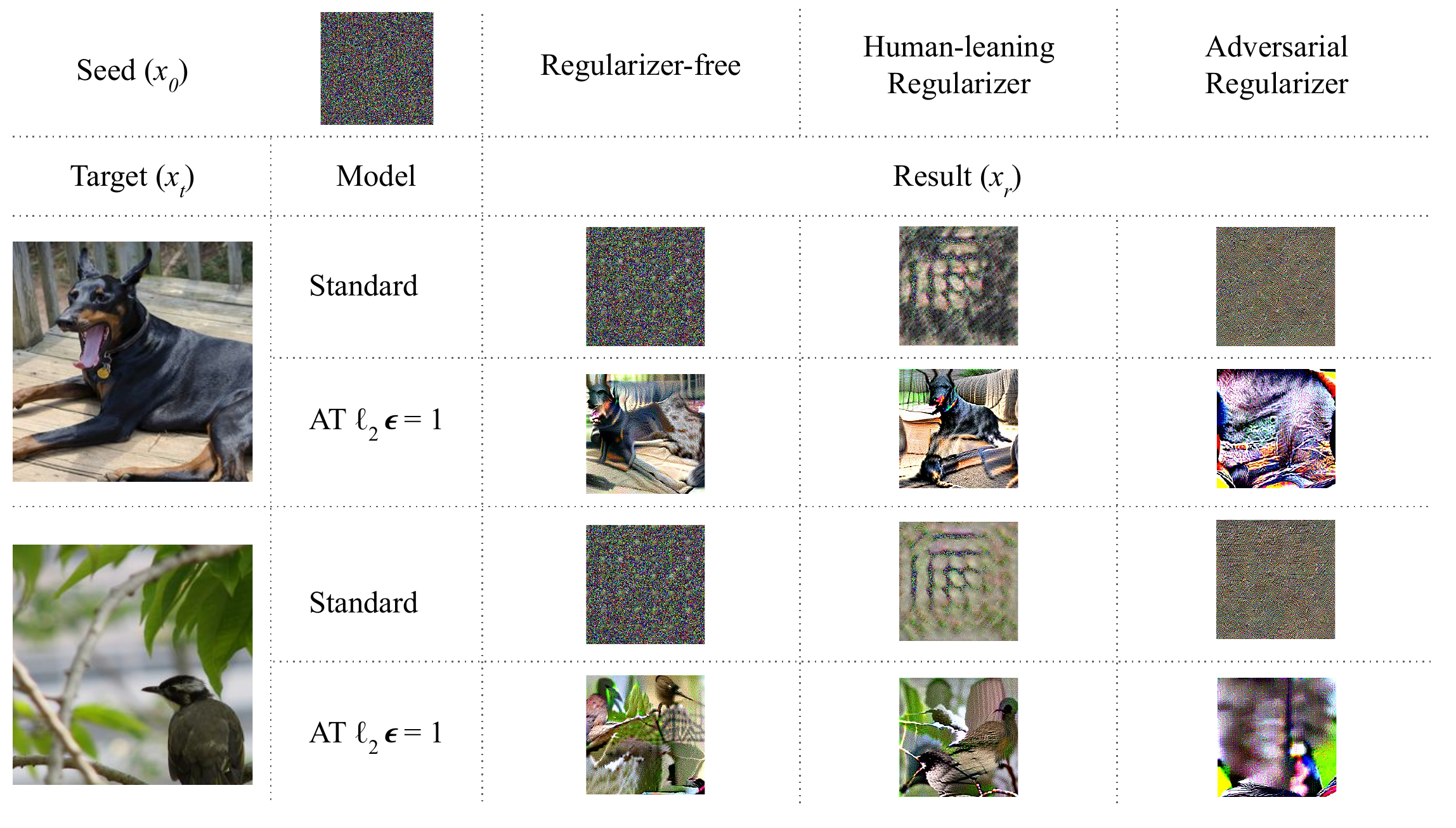}
        
\caption{ \textbf{[Representation Inversion for different kinds of $\mathcal{R}$; For ImageNet trained ResNet50]} For the standard ResNet50 (trained using cross-entropy loss), with regularizer-free and adversarial inversion, $x_r$ looks perceptually much closer to $x_0$ than $x_t$, even though from the model's point of view, $x_r$ and $x_t$ are the same. However, with the human-leaning regularizer, we see that $x_r$ contains some information like color patterns of $x_t$. For adversarially robust ResNet50~\cite{madry2019deep,salman2020adversarially} even though regularizer-free and human-leaning inversions look perceptually similar to $x_t$, for the adversarial regularizer even these models produce $x_r$ that looks nothing like $x_t$. Images are generated by starting from $x_0$ and solving Eq~\ref{eq:inversion} with different kinds of regularizers.
}
\label{fig:inversion_examples}
\end{figure*}

We seek to make sense of these confusing earlier results, and thereby to better understand alignment. We show that when we evaluate alignment of DNNs' invariances and human perception using \iris generated using different loss functions, we can arrive at very different conclusions. For example, Fig~\ref{fig:inversion_examples} shows how visual similarity of \iris~can vary massively across different categories of losses.

We group existing \iri~generation processes into two broad categories: \textit{regularizer-free} (as in~\cite{jenelle2019metamers}), where the goal is to find an \iri~without any additional constraints; and \textit{human-leaning} (as in~\cite{olah2017feature, olah2020zoom, mahendran2014understanding}), where the goal is to find an \iri~that is also visually human-comprehensible.
Additionally, we propose and explore a new (third) broad category, \textit{adversarial}, where the goal is to find an \iri~that is visually (from a human perception perspective) far apart from the target input.

We find that compared to the regularizer-free \iri~generation approach, the human-leaning IRI generation approach applies strong constraints on the kind of \iris~generated and thus limits the ability to freely explore the large space of possible \iris. On the other hand, our proposed adversarial approach shows that in the worst case, all models have close to zero alignment, suggesting that there is scope for improvement in designing models that have human-like invariances (as shown in Fig~\ref{fig:inversion_examples} and Table~\ref{tab:regularizer_results}). 
Based on this evidence, we argue that in order to have meaningful comparisons between models, one should measure alignment using the regularizer-free loss for \iri~generation.

Many prior works do not formally define a measure that can quantify alignment with human perception beyond relying on visual inspection of the images by the authors~\cite{olah2017feature,olah2020zoom,mahendran2014understanding,engstrom2019adversarial}. We show how alignment can be quantified reliably by designing simple visual perception tests that can be crowdsourced, \ie~used in human surveys. We also show how one can leverage widely used measures of perceptual distance~\cite{zhang2018perceptual} to automate our human surveys, which allows us to obtain insights at a scale not possible in previous works.

Next, inspired by the prior works that suggest that changes in the model training pipeline (as in training adversarially robust DNNs~\cite{engstrom2019adversarial,kaur2019perceptually}) can lead to human-like invariant representations, we conduct an in-depth investigation to understand which parts of the deep learning pipeline are critical in helping DNNs better learn human-like invariances. 
We find that certain choices in the deep learning pipeline can significantly help learn representation that have human-like invariances. For example, we show that residual architectures (\eg, ResNets~\cite{he2016deep}), when trained with a self-supervised contrastive loss (\eg, SimCLR~\cite{chen2020simple}), using $\ell_2$ ball adversarial data augmentations (\eg, as in RoCL~\cite{kim2020adversarial}); the learned representations -- while typically having lower accuracies than their fully supervised counterparts -- have higher alignment of invariances with human perception. We highlight the following contributions:

\begin{itemize}[leftmargin=*]
    \item We show how different losses used for generating \iris~lead to different conclusions about a model's shared invariances with human perception, thus leading to seemingly contradictory findings in prior works. 
    \item We propose an adversarial \iri~generation loss, using which we show empirically that we can almost always discover invariances of DNNs that do not align with human perception, thus suggesting that there is scope to design better mechanisms to learn representations that are more aligned with human perception.
    \item We conduct an in-depth study of how loss functions, architectures, data augmentations and training paradigms lead to learning human-like shared invariances. 
\end{itemize}

\begin{table*}[t]
\captionsetup{font=footnotesize}
\begin{center}
\begin{small}
\begin{sc}
\begin{tabular}{c|c|ccccc}
\hline

\multicolumn{7}{c}{CIFAR10} \\
\hline

 & \multirow{2}{*}{Model} & 
\multirow{2}{*}{\begin{tabular}[c]{@{}c@{}}Human\\2AFC\end{tabular}} &
\multirow{2}{*}{\begin{tabular}[c]{@{}c@{}}LPIPS\\2AFC\end{tabular}} &
\multirow{2}{*}{\begin{tabular}[c]{@{}c@{}}Human\\Clustering\end{tabular}} &
\multirow{2}{*}{} &
\multirow{2}{*}{\begin{tabular}[c]{@{}c@{}}LPIPS\\Clustering\end{tabular}} \\
 & & & & & & \\
\hline

\multirow{4}{*}{\begin{tabular}[c]{@{}c@{}}AT\\$\ell_2$ \\ $\epsilon = 1$\end{tabular}} 
 & ResNet18 & $96.00_{\pm 2.55}$ & $87.25_{\pm 9.52}$ & $97.48_{\pm 1.80}$ &  & $88.13_{\pm 6.57}$ \\
 & VGG16 & $38.83_{\pm 7.59}$ & $4.00_{\pm 3.86}$ & $55.39_{\pm 5.63}$ &  & $46.09_{\pm 3.84}$ \\
 & InceptionV3 & $82.00_{\pm 8.44}$ & $54.12_{\pm 19.23}$ & $84.47_{\pm 6.32}$ &  & $74.87_{\pm 6.74}$ \\
 & Densenet121 & $98.67_{\pm 0.24}$ & $91.75_{\pm 8.2}$ & $97.64_{\pm 2.08}$ &  & $91.92_{\pm 6.13}$ \\
 
\hline

\multirow{4}{*}{\begin{tabular}[c]{@{}c@{}}Standard\end{tabular}}
 & ResNet18 & $0.17_{\pm 0.24}$ & $0.0_{\pm 0.0}$ & $38.55_{\pm 1.19}$ &  & $35.35_{\pm 3.27}$ \\
 & VGG16 & $0.17_{\pm 0.24}$ & $0.0_{\pm 0.0}$ & $33.84_{\pm 2.70}$ &  & $32.58_{\pm 1.04}$ \\
 & InceptionV3 & $0.17_{\pm 0.24}$ & $0.38_{\pm 0.41}$ & $38.38_{\pm 4.06}$ &  & $36.62_{\pm 3.08}$ \\
 & Densenet121 & $9.83_{\pm 9.97}$ & $0.12_{\pm 0.22}$ & $42.42_{\pm 5.02}$ &  & $37.12_{\pm 3.54}$ \\

\hline

\multicolumn{7}{c}{IMAGENET} \\
\hline

 & \multirow{3}{*}{Model} & 
\multirow{3}{*}{\begin{tabular}[c]{@{}c@{}}Human\\2AFC\end{tabular}} &
\multirow{3}{*}{\begin{tabular}[c]{@{}c@{}}LPIPS\\2AFC\end{tabular}} &
\multirow{3}{*}{\begin{tabular}[c]{@{}c@{}}Human\\Clustering\end{tabular}} &
\multirow{3}{*}{\begin{tabular}[c]{@{}c@{}}Human\\Clustering\\Hard\end{tabular}} &
\multirow{3}{*}{\begin{tabular}[c]{@{}c@{}}LPIPS\\Clustering\end{tabular}} \\
 & & & & \\
 & & & & \\

\hline

\multirow{3}{*}{\begin{tabular}[c]{@{}c@{}}AT\\$\ell_2$ \\ $\epsilon = 3$\end{tabular}} 
 & ResNet18 & $93.17_{\pm 5.95}$ & $53.37_{\pm 20.19}$ & $96.00_{\pm 3.59}$ & $87.75_{\pm 7.60}$ & $65.28_{\pm 10.58}$ \\
 & ResNet50 & $99.50_{\pm 0.00}$ & $53.63_{\pm 20.64}$ & $99.49_{\pm 0.71}$ & $97.06_{\pm 3.47}$ & $71.21_{\pm 9.93}$ \\
 & VGG16 & $95.50_{\pm 2.12}$ & $59.38_{\pm 21.48}$ & $91.75_{\pm 5.22}$ & $90.69_{\pm 3.13}$ & $70.33_{\pm 9.78}$  \\
 
\hline

\multirow{4}{*}{\begin{tabular}[c]{@{}c@{}}Standard\end{tabular}}
 & ResNet18 & $0.00_{\pm 0.00}$ & $1.12_{\pm 1.67}$ & $33.33_{\pm 0.00}$ & - & $34.60_{\pm 0.56}$ \\
 & ResNet50 & $5.33_{\pm 7.54}$ & $0.38_{\pm 0.41}$ & $38.38_{\pm 2.53}$ & - & $35.35_{\pm 0.62}$  \\
 & VGG16 & $0.00_{\pm 0.00}$ & $0.00_{\pm 0.00}$ & $33.96_{\pm 2.00}$ & - & $34.47_{\pm 1.49}$ \\
 
\hline

\end{tabular}
\end{sc}
\end{small}
\end{center}

\caption{\textbf{[CIFAR10 and ImageNet Surveys To Confirm Efficacy of LPIPS]} We use LPIPS to simulate a human in both 2AFC and Clustering setups described in Section~\ref{sec:g_human} and compare it with AMT worker's responses. A value close to $33\%$ for clustering means random assignment and indicates no alignment. We see that LPIPS and humans rank models similarly in both 2AFC and clustering setups, thus showing that LPIPS is a reliable proxy for judging perceptual similarity of \iris. These experiments were conducted on \iris~generated using regularizer-free loss in Eq~\ref{eq:inversion}. The variance reported for LPIPS is for different backbone networks that are available for LPIPS.
}
\vspace{-3mm}
\label{tab:survey_results}
\end{table*}

\section{Measuring Shared Invariance with Human Perception}\label{sec:measuring_alignment}

Measuring the extent to which invariances learned by DNNs are shared by humans is a two step process. We first  generate \iris, \ie, inputs that are mapped to identical representations by the DNN. \iris~give us an estimate about the invariances of the DNN. Then, we assess if these inputs are also considered identical by humans. More concretely, if invariances of a given DNN ($g_{\text{model}}$) are shared by humans ($g_{\text{human}}$) on a set of $n$ $d$-dimensional samples $X \in \mathbb{R}^{n \times d}$, then: 

\begin{gather*}
    g_{\text{human}}(X^i) \approx g_{\text{human}}(X^j) \, \forall \, (X^i, X^j) \in \mathcal{S} \times \mathcal{S} \,\,;\,\, \\\mathcal{S} = \{X\} \cup \{X^i \,\,|\,\, g_{model}(X^i) \approx g_{model}(X)\}.
\end{gather*}
$\mathcal{S}$ denotes the \iris~for $g_{\text{model}}$. There are three major challenges here:
\begin{itemize}
    \item Access to representations in the brain, \ie, $g_{\text{human}}$ is not available.
    \item Due to the highly non-linear nature of DNNs, $\mathcal{S}$ can be very hard to obtain.
    \item The %
    fine-grained input space implies %
    very many inputs $n$, making the choice of $X$ hard.
\end{itemize}

We address each of these below. We also show how prior works that do not directly engage with these points can miss important issues in their conclusions about shared invariances of DNNs and humans.

\begin{figure*}[t!]
    \centering
    \begin{subfigure}[b]{0.31\textwidth}
        \raisebox{0.5\height}{\includegraphics[width=1\textwidth]{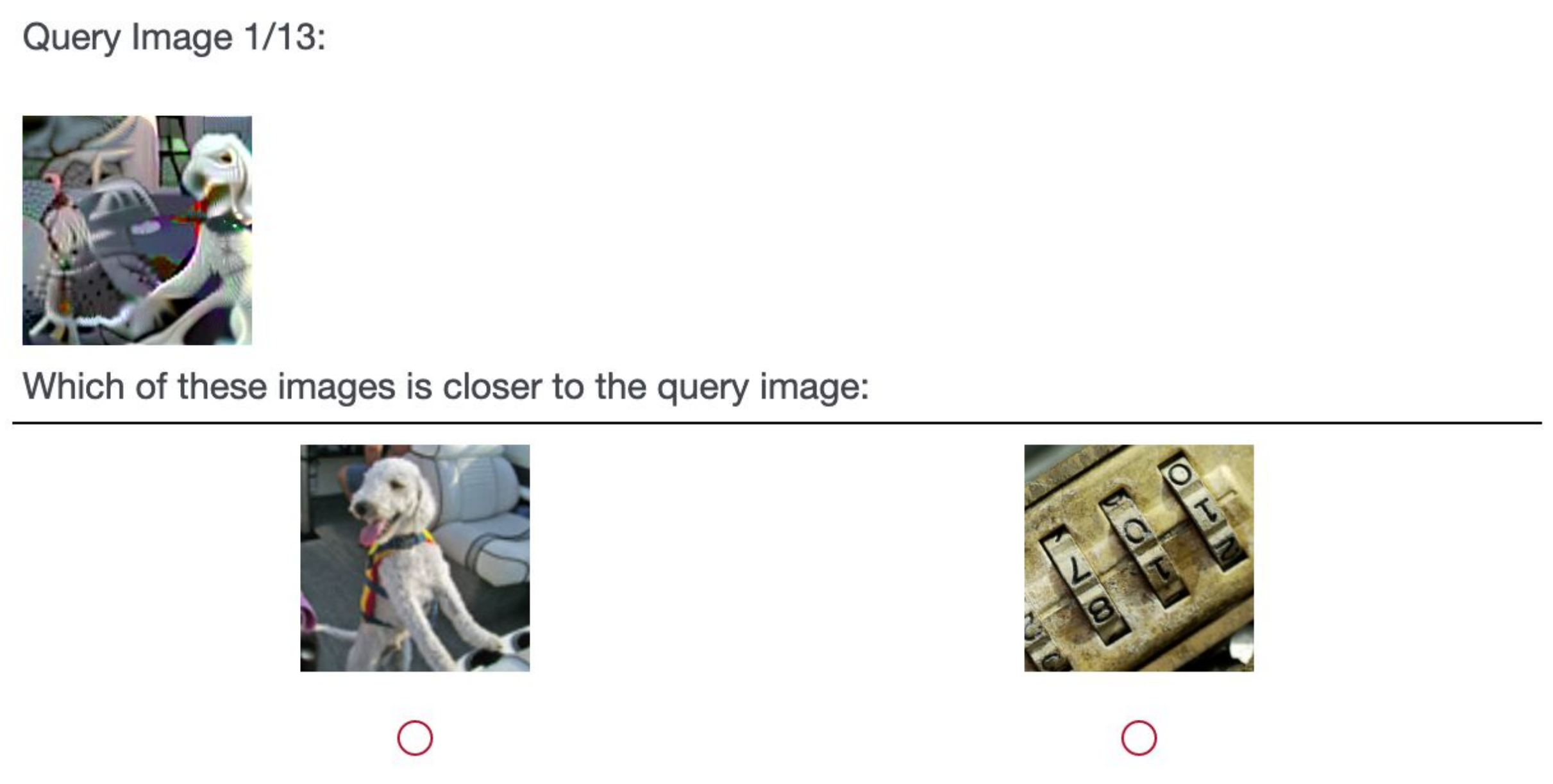}}
        \caption{2AFC}
        \label{fig:2afc_survey}
    \end{subfigure}
    \begin{subfigure}[b]{0.3\textwidth}
        \includegraphics[width=0.75\textwidth]{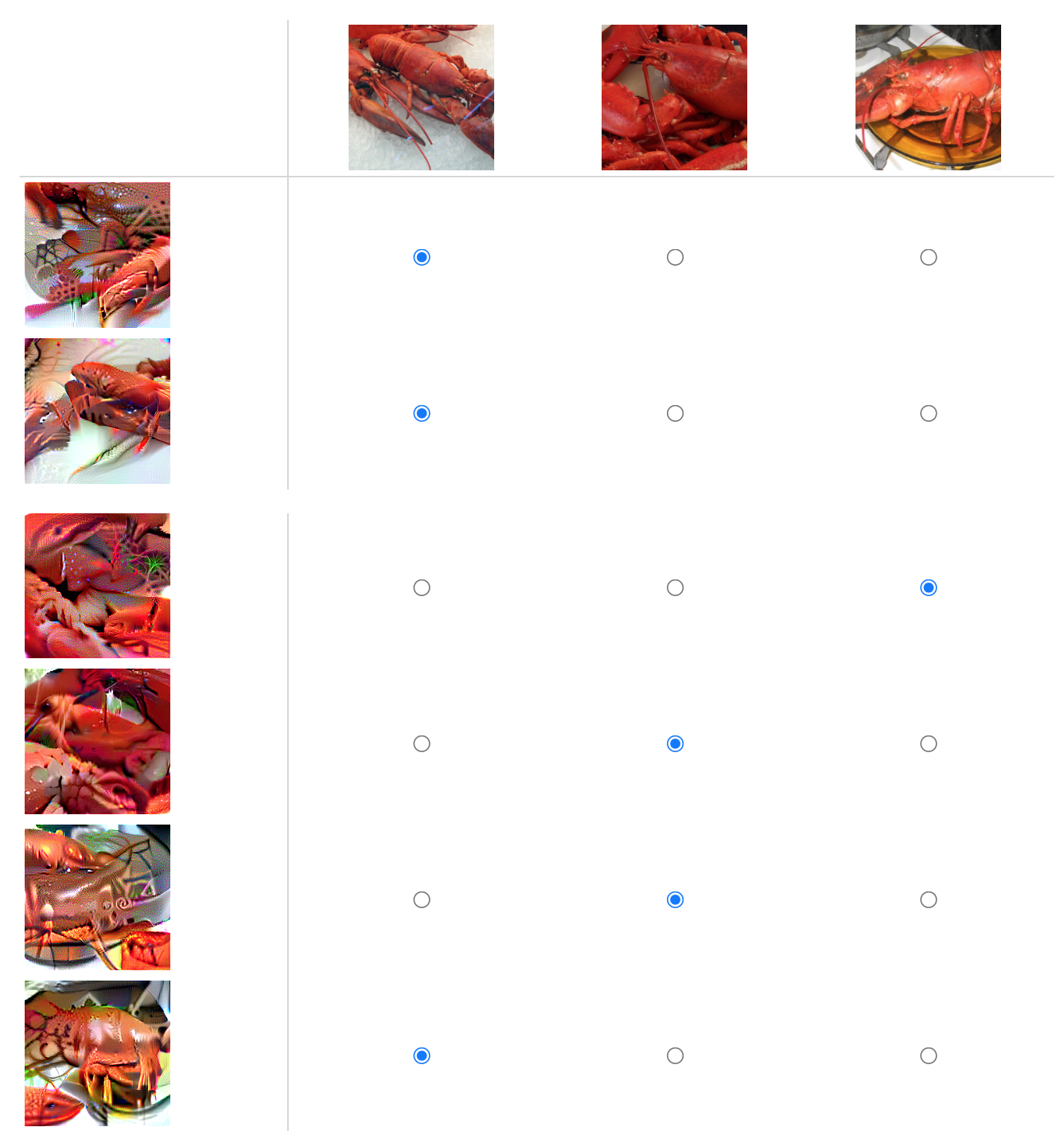}
        \caption{Hard ImageNet Clustering}
        \label{fig:imagenet_clustering_hard}
    \end{subfigure}
    \begin{subfigure}[b]{0.3\textwidth}
        \includegraphics[width=1.15\textwidth]{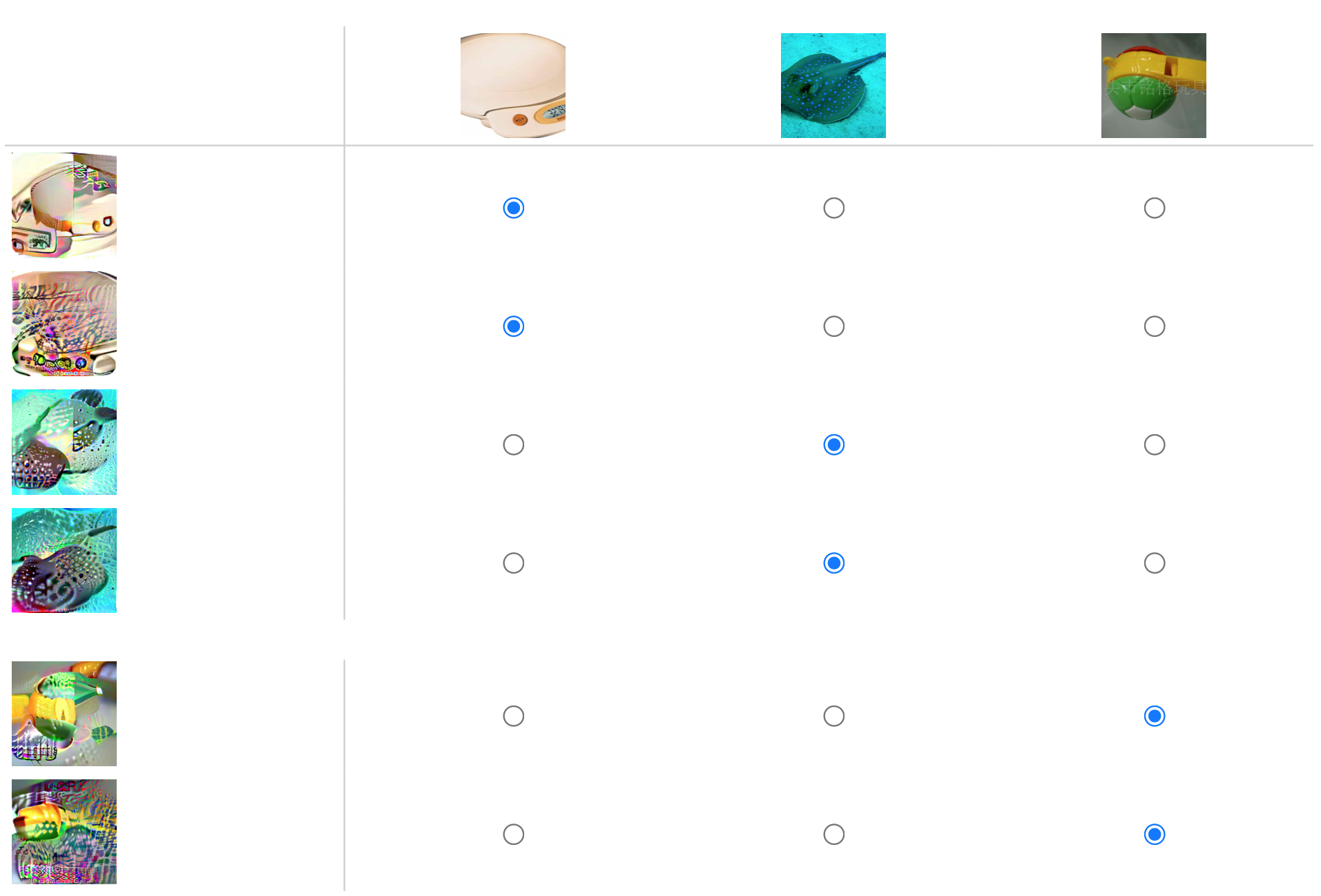}
        \caption{Random ImageNet Clustering}
        \label{fig:imagenet_clustering_easy}
    \end{subfigure}
\caption{ \textbf{[Survey Prompts for AMT workers]} In the 2AFC (left) setting we ask the annotator to choose which of the two images ($x_t$ or $x_0$) is perceptually closer to the query image ($x_r$). 
In the clustering setting (center and right) we show 3 images from the dataset (target images, $x_t$) in the columns and for each of these, we generate $x_{r_1} \in \mathcal{S}_{x_t}$ and $x_{r_2} \in \mathcal{S}_{x_t}$. 
Each of these is shown across the rows. The task here is to match each image on the row with the corresponding target image on the column. 
}
\label{fig:survey_prompts}
\end{figure*}

\subsection{Approximating $g_{\text{human}}$}\label{sec:g_human}

Assuming we have a set of images with identical representations ($\mathcal{S}$; how we obtain this is discussed in Section~\ref{sec:generating_iris}), we must check if humans also perceive these images to be identical. The extent to which humans think this set of images is identical defines how aligned the invariances learned by the DNN are with human perception. In prior works this has been done by either eyeballing \iris~\cite{engstrom2019adversarial} or by asking annotators to assign class labels to \iris~\cite{jenelle2019metamers}; both approaches do not scale well. Additionally, assigning class labels to \iris~limits $X$ to being samples from a data distribution containing human-recognizable images (\ie, X cannot be sampled from any arbitrary distribution) with only a few annotations (\eg, asking annotators to assign one class label out of 1000 ImageNet classes is not feasible). To address the issues of scalability and class labels, we propose the following as a measure of alignment between DNN and human invariances:

\begin{gather}\label{eq:alignment_measure}
    \text{Alignment} = \frac{|\mathcal{A}|}{\sum_{x_t \in X}|\mathcal{S}_{x_t}|}, \text{where}
\end{gather}
\begin{gather*}
\,\,\, \mathcal{A} = \{x_{r_i} | \,\,\, ||g_{\text{human}}(x_t) - g_{\text{human}}(x_{r_i}) || < \\ || g_{\text{human}}(x_0) - g_{\text{human}}(x_{r_i}) || \,\,\, \forall x_t \in X, x_{r_i} \in S_{x_t} \}, \\
    S_{x_t} = \{x_{r_i} | g_{\text{model}}(x_{r_i}) \approx g_{\text{model}}(x_t) \,\,\, \forall x_t \in X \}, 
\end{gather*}

where $x_0$ is the starting point for Eq~\ref{eq:inversion} sampled from $\mathcal{N}(0,1)$. In Section~\ref{sec:eval_and_takeaways} we see how alignment is robust to the choice of $x_0$. By directly looking for perceptual similarity of \iris~(captured by $\mathcal{A}$), we get past the issue of assigning class labels to \iris. The comparison used to generate $\mathcal{A}$ is referred to as the 2 alternative forced choice test (2AFC) which is commonly used to assess sensitivity of humans to stimuli~\cite{fechner1948elements}. In order to compute $\mathcal{A}$, we estimate perceptual distance $d(x_i, x_j) = || g_{\text{human}}(x_i) - g_{\text{human}}(x_j) ||$ between two inputs. Ideally, we would like to measure $d(x_i, x_j)$ by directly asking for human annotations, however, this approach is expensive and does not scale when we wish to evaluate many models. To address scalability, we use LPIPS~\cite{zhang2018perceptual} which is a commonly used measure for perceptual distance and thus can be used to approximate $d(x_i, x_j)$~\footnote{For all evaluations we report the average over 4 different backbones used to calculate LPIPS including the finetuned weights released by the authors. More details in Appendix~\ref{sec:g_human_appendix}}. While LPIPS is by no means a perfect approximation, it allows us to gain insights at a scale not possible in prior works.  

To ensure the efficacy of LPIPS as a proxy for human judgements, we deploy two types of surveys on Amazon Mechanical Turk (AMT) to also elicit human similarity judgements. %
Prompts for these surveys are shown in Fig.~\ref{fig:survey_prompts}. We received approval from the Ethical Review Board of our institute for this survey. Each survey consists of 100 images plus some attention checks to ensure the validity of our responses. The survey was estimated to take 30 minutes (even though on average our annotators took less than 20 minutes), and we paid each worker 7.5 USD per survey. 

\xhdr{Clustering}  In this setting, we ask humans to match the \iris~($x_{r_i}$) on the row to the most perceptually similar image ($x_t$) on the column (each row can only be matched to one column). A prompt for this type of a task is shown in Fig.~\ref{fig:imagenet_clustering_hard} \& \ref{fig:imagenet_clustering_easy}. With these responses, we calculate a quantitative measure of alignment by measuring the fraction of $x_{r_i}$ that were correctly matched to their respective $x_t$. For ImageNet, we observed that a random draw of three images (\eg, Fig.~\ref{fig:imagenet_clustering_easy}) can often be easy to match to based on how different the drawn images ($x_t$) are. Thus, we additionally construct a ``hard'' version of this task by ensuring that the three images are very ``similar'' (as shown in Fig.~\ref{fig:imagenet_clustering_hard}). We leverage human annotations of ImageNet-HSJ~\cite{roads21enriching} to draw these similar images. More details can be found in Appendix~\ref{sec:human_alignment_appendix}.

\xhdr{2AFC} This is the exact test used to generate $\mathcal{A}$. In this setting we show the annotator a reconstructed image ($x_r$) and ask them to match it to one of the two images shown in the options. The images shown in the options are the seed ($x_0$, \ie, starting value of $x$ in Eq.~\ref{eq:inversion}) and the original image ($x_t$). Since $x_r$ and $x_t$ are \iris~for the model (by construction), alignment would imply humans also perceive $x_r$ and $x_t$ similarly. See Fig.~\ref{fig:2afc_survey} for an example of this type of survey.

\subsection{Generating \iris}\label{sec:generating_iris}
Even if we assume a finite sampled set $X \sim \mathcal{D}$ (discussed in Section~\ref{sec:choosing_x}), there can be many samples in $\mathcal{S}$ due to the highly non-linear nature of DNNs. However, we draw on the insight that there is often some structure to the set of \iris, that is heavily dependent on the  IRI generation process. Prior work on understanding shared invariance between DNNs and humans~\cite[e.g.,][]{engstrom2019adversarial, jenelle2019metamers} has used representation inversion~\citep{mahendran2014understanding} to generate \iris. However, \iris~generated this way depend heavily on the loss function used in representation inversion, as demonstrated by \cite{olah2017feature}. Fig.~\ref{fig:inversion_examples} shows how different loss functions can lead to very different looking \iris. We group these losses previously used in the literature to generate \iris~into two broad types: \textbf{\textit{regularizer-free}} (used by \cite{engstrom2019adversarial,jenelle2019metamers}), and \textbf{\textit{human-leaning}} (used by many works on interpretability of DNNs including \cite{mahendran2014understanding,olah2017feature,olah2020zoom,Mordvintsev2015inceptionism,Nguyen2015deep}). We also explore a third kind of \textbf{\textit{adversarial}} regularizer, that aims to generate \textit{controversial stimuli}~\cite{golan2020controversial} between a DNN and a human.

Representation inversion is the task of starting with a random seed image $x_0$ to reconstruct a given image $x_t \in X$ from its representation $g(x_t)$ where $g(\cdot)$ is the trained DNN. The reconstructed image ($x_r$) is same as $x_t$ from the DNN's point of view, \ie, $g(x_t) \approx g(x_r)$. This is achieved by performing gradient descent on $x_0$ (in our experiments we use SGD with a learning rate of $0.1$) to minimize a loss of the following general form:

\begin{equation}
\label{eq:inversion}
    \mathcal{L}_x = \frac{|| g(x_t) - g(x) ||_2}{||g(x_t)||_2} + \lambda * \mathcal{R}(x)
\end{equation}

where $\lambda$ is an appropriate scaling constant for regularizer $\mathcal{R}$. All of these reconstructions induce representations in the DNN that are very similar to the given image ($x_t$), as measured using $\ell_2$ norm. Depending on the choice of seed $x_0$ and the choice of $\mathcal{R}$, we get different reconstructions of $x_t$ thus giving us a set of inputs $\{x_t, x_{r_1},...,x_{r_k}\}$ that are all mapped to similar representations by $g(\cdot)$. Doing this for all $x_t \in X$, we get the \iris~, $\mathcal{S} = \{X, X^{r_1},...,X^{r_k}\}$. 

In practice we find that the seed $x_0$ does not have any significant impact on the measurement of shared invariance. However, the choice of $\mathcal{R}$ \emph{does} significantly impact the invariance measurement (as also noted by~\cite{olah2017feature}). We identify the following distinct categories of \iris~based on the choice of $\mathcal{R}$.

\textbf{\textit{Regularizer-free.}} These methods do not use a regularizer, \ie, $\mathcal{R}(x) = 0$.

\textbf{\textit{human-leaning regularizer.}} This kind of a regularizer purposefully puts constraints on $x$ such that the reconstruction has some ``meaningful'' features. \cite{mahendran2014understanding} use $\mathcal{R}(x) = TV(x) + ||x||_p$ where $TV$ is the total variation in the image. Intuitively this penalizes high frequency features and smoothens the image to make it look more like natural images. \cite{Nguyen2015deep} achieve a similar kind of high frequency penalization by blurring $x$ before each optimization step. We combine both these frequency-based regularizers with pre-conditioning in the Fourier domain~\cite{olah2017feature} and robustness to small transformations~\cite{Nguyen2015deep}. More details can be found in Appendix~\ref{sec:regulziers_appendix}. Intuitively a regularizer from this category generates \iris~that have been ``biased'' to look meaningful to humans.

\textbf{\textit{Adversarial regularizer.}} We propose a new regularizer to generate \iris~while intentionally making them look \emph{perceptually dissimilar} from the target, \ie, $\mathcal{R} = - || g_{\text{human}}(x_t) - g_{\text{human}}(x) ||$ (negative sign since we want to maximize perceptual distance between $x$ and $x_t$). 
We leverage LPIPS (Learned Perceptual Image Patch Similarity)~\cite{zhang2018perceptual}, a widely used \textit{perceptual distance} measure, to approximate $|| g_{\text{human}}(x_t) - g_{\text{human}}(x) ||$. LPIPS uses initial layers of an ImageNet trained model (finetuned on a dataset of human similarity judgements) to approximate perceptual distance between images which makes it differentiable and thus can be easily plugged into Eq.~\ref{eq:inversion}. Thus, the regularizer used is $\mathcal{R}(x) = -\, \text{LPIPS}(x, x_t)$. \iris~generated using this regularizer can be thought of as \textit{controversial stimuli}~\cite{golan2020controversial} -- they're similar from the DNN's perspective, but distinct from a human's perspective.

\subsection{Choice of inputs $X$}\label{sec:choosing_x}
In order to overcome the challenge of choosing $X$, we assume $X$ to be sampled from a given data distribution $\mathcal{D}$. In our experiments, we try out many different distributions, including the training data distribution and random noise distributions, and find that takeaways about a alignment of model's invariances with humans do not depend heavily on the choice of $\mathcal{D}$. Some examples of $X$ sampled from the data distribution and noise distributions (two random Gaussian distributions, $\mathcal{N}(0, 1)$ and $\mathcal{N}(0.5, 2)$), along with the corresponding \iris~are shown in Fig.~\ref{fig:in_vs_ood_reconstructions}, Appendix~\ref{sec:choosing_x_appendix}. Interestingly, the human-leaning regularizer, which explicitly tries to remove high-frequency features from $x_r$ fails to reconstruct an $x_t$ that itself consists of high-frequency features.

\begin{table*}[t]
\captionsetup{font=footnotesize}
\begin{center}
\begin{small}
\begin{sc}
\begin{tabular}{c|c|ccc|cc}
\hline

\multicolumn{7}{c}{CIFAR10} \\
\hline

\multirow{3}{*}{Training} & \multirow{2}{*}{Model} & 
\multicolumn{3}{c|}{\begin{tabular}[c]{@{}c@{}}Alignment\end{tabular}}&
\multirow{3}{*}{\begin{tabular}[c]{@{}c@{}}Clean\\Acc.\end{tabular}} & \multirow{3}{*}{\begin{tabular}[c]{@{}c@{}}Robust\\Acc.\end{tabular}} \\ \cline{3-5}
 & & \begin{tabular}[c]{@{}c@{}}Reg.-\end{tabular} &
\begin{tabular}[c]{@{}c@{}}Human-\end{tabular} &
\begin{tabular}[c]{@{}c@{}}Adver-\end{tabular} & & \\
 & & \begin{tabular}[c]{@{}c@{}}Free\end{tabular} & \begin{tabular}[c]{@{}c@{}}Aligned\end{tabular} & \begin{tabular}[c]{@{}c@{}}sarial\end{tabular} & & \\
\hline

\multirow{4}{*}{\begin{tabular}[c]{@{}c@{}}AT\\$\ell_2, \epsilon = 1$\end{tabular}} 
 & ResNet18 & $63.25_{\pm 26.23}$ & $79.00_{\pm 21.94}$ & $0.33_{\pm 0.47}$ & $80.77$ & $50.92$ \\
 & VGG16 & $0.25_{\pm 0.43}$ & $41.41_{\pm 16.74}$ & $1.00_{\pm 1.41}$ & $79.84$ & $48.36$ \\
 & InceptionV3 & $23.25_{\pm 25.56}$ & $64.75_{\pm 24.17}$ & $3.00_{\pm 4.24}$ & $81.57$ & $51.02$ \\
 & Densenet121 & $82.75_{\pm 20.07}$ & $86.25_{\pm 14.50}$ & $1.33_{\pm 1.89}$ & $83.22$ & $52.86$ \\
 
\hline

\multirow{4}{*}{\begin{tabular}[c]{@{}c@{}}Standard\end{tabular}}
 & ResNet18 & $0.00_{\pm 0.00}$ & $21.09_{\pm 13.51}$ & $1.33_{\pm 1.89}$ & $94.94$ & $0.00$ \\
 & VGG16 & $0.00_{\pm 0.00}$ & $21.88_{\pm 14.82}$ & $0.00_{\pm 0.00}$ & $93.63$ & $0.00$ \\
 & InceptionV3 & $0.00_{\pm 0.00}$ & $21.88_{\pm 17.54}$ & $0.33_{\pm 0.47}$ & $94.59$ & $0.00$ \\
 & Densenet121 & $0.00_{\pm 0.00}$ & $26.56_{\pm 16.90}$ & $0.00_{\pm 0.00}$ & $95.30$ & $0.00$ \\

\hline

\multicolumn{7}{c}{IMAGENET} \\
\hline

\multirow{3}{*}{Training} & \multirow{2}{*}{Model} & 
\multicolumn{3}{c|}{\begin{tabular}[c]{@{}c@{}}Alignment\end{tabular}}&
\multirow{3}{*}{\begin{tabular}[c]{@{}c@{}}Clean\\Acc.\end{tabular}} & \multirow{3}{*}{\begin{tabular}[c]{@{}c@{}}Robust\\Acc.\end{tabular}} \\ \cline{3-5}
 & & \begin{tabular}[c]{@{}c@{}}Reg.-\end{tabular} &
\begin{tabular}[c]{@{}c@{}}Human-\end{tabular} &
\begin{tabular}[c]{@{}c@{}}Adver-\end{tabular} & & \\
 & & \begin{tabular}[c]{@{}c@{}}Free\end{tabular} & \begin{tabular}[c]{@{}c@{}}Aligned\end{tabular} & \begin{tabular}[c]{@{}c@{}}sarial\end{tabular} & & \\
\hline

\multirow{4}{*}{\begin{tabular}[c]{@{}c@{}}AT\\$\ell_2, \epsilon = 3$\end{tabular}}  
 & ResNet18 & $42.00_{\pm 38.33}$ & $46.75_{\pm 39.37}$ & $0.33_{\pm 0.47}$ & $53.12$ & $31.02$ \\
 & ResNet50 & $51.00_{\pm 34.89}$ & $45.75_{\pm 37.39}$ & $14.00_{\pm 3.74}$ & $62.83$ & $38.84$  \\
 & VGG16 & $55.50_{\pm 34.14}$ & $55.50_{\pm 38.29}$ & $11.00_{\pm 3.74}$ & $56.79$ & $34.46$  \\
 
\hline

\multirow{4}{*}{\begin{tabular}[c]{@{}c@{}}Standard\end{tabular}}
 & ResNet18 & $0.00_{\pm 0.00}$ & $17.00_{\pm 28.30}$ & $0.00_{\pm 0.00}$ & $69.76$ & $0.01$ \\
 & ResNet50 & $0.00_{\pm 0.00}$ & $16.25_{\pm 26.42}$ & $0.00_{\pm 0.00}$ & $76.13$ & $0.00$  \\
 & VGG16 & $0.00_{\pm 0.00}$ & $0.00_{\pm 0.00}$ & $0.00_{\pm 0.00}$ & $73.36$ &  $0.16$ \\
 
\hline

\end{tabular}
\end{sc}
\end{small}
\end{center}

\caption{\textbf{[CIFAR10 and ImageNet Model Alignment Results for Different Regularizers]} For different regularizers, we see that ranking of models can look very different. For example, for Adversarially Trained (AT) Resnet18 vs InceptionV3 on CIFAR10, we see that reguarizer-free inversion leads to Resnet18 being significantly more aligned, but the trend is much less pronounced for the human-leaning regularizer. We also find that alignment can vary quite a bit between different architectures -- all of which achieve similar clean and robust accuracies. 
}
\vspace{-3mm}
\label{tab:regularizer_results}
\end{table*}

\subsection{Evaluation and Takeaways}\label{sec:eval_and_takeaways}

For each model, we randomly picked 100 images from the data distribution along with a seed image with random pixel values. For each of the 100 images, we do representation inversion using one regularizer each from \textit{regularizer-free}, \textit{human-leaning}, and \textit{adversarial}. 

\xhdr{Reliability of using LPIPS} Table~\ref{tab:survey_results} shows the results for the surveys conducted with AMT workers~\footnote{This was conducted only using \iris~from regularizer-free inversion.}. Each survey was completed by 3 workers. For a well aligned model, the scores under \texttt{2AFC} and \texttt{Clustering} should be close to 1, while for a non-aligned model scores under \texttt{2AFC} should be close to 0, and scores under \texttt{Clustering} should be close to a random guess (\ie, about $33\%$). We see that LPIPS (with different backbone nets, e.g., AlexNet, VGG) orders models similar to human annotators for both the survey setups, thus showing that it's a reliable proxy.

\xhdr{Reliability of Human Annotators} In Table~\ref{tab:survey_results}, we make three major observations: 1) variance between different annotators is very low; 2) scores under \texttt{Human 2AFC} and \texttt{Human Clustering} order different models similarly; and finally, 3) even though accuracy drops for the ``hard'' version of ImageNet task, the relative ordering of models remains the same. These observations indicate that alignment can be reliably measured by generating \iris~and does not depend on bias in annotators. Note that AMT experiments were only performed on \iris~generated using the regularizer-free loss in Eq~\ref{eq:inversion}.

\xhdr{Impact of regularizer} Table~\ref{tab:regularizer_results} shows the results of Alignment (Eq~\ref{eq:alignment_measure}) for different regularizers for \iri~generation. We evaluated multiple architectures of both standard and adversarially trained~\cite{madry2019deep} CIFAR10 and ImageNet models. We find that under different types of regularizers, the alignment of models can look very different. We also see that adversarial regularizer makes aligment bad for almost all models, thus showing that for the worst pick of \iris~the alignment between learned invariances and human invariances has a lot of room for improvement. Conversely, the human-leaning regularizer overestimates the alignment.

\xhdr{Impact of $X$} In the case of OOD targets ($x_t$) we see that humans are still able to faithfully judge similarity, yielding the same ranking of models as in-distribution targets. Some results for human judgements about similarity of \iris~for out of distribution samples are shown in Table~\ref{tab:in_vs_ood_survey}, Appendix~\ref{sec:choosing_x_appendix}. As seen in Fig~\ref{fig:in_vs_ood_reconstructions} (Appendix~\ref{sec:choosing_x_appendix}), human-leaning regularizer does not work well for reconstructing noisy targets. This is because such regularizers explicitly remove high-frequency features from reconstructions~\cite{olah2017feature} and thus struggle to meaningfully reconstruct targets that contain high-frequency features. Hence, all results in Table~\ref{tab:in_vs_ood_survey}, Appendix~\ref{sec:choosing_x_appendix} are reported on \iris~generated using regularizer-free loss.

\xhdr{Impact of $x_0$} We repeat some of the experiments with other starting points for Eq~\ref{eq:inversion} and find that results are generally not sensitive to the choice of $x_0$. Results are included in Appendix~\ref{sec:role_of_seed_appendix}.

    \begin{figure*}[t]
        \centering
        \begin{subfigure}[b]{0.28\textwidth}
            \includegraphics[width=\textwidth]{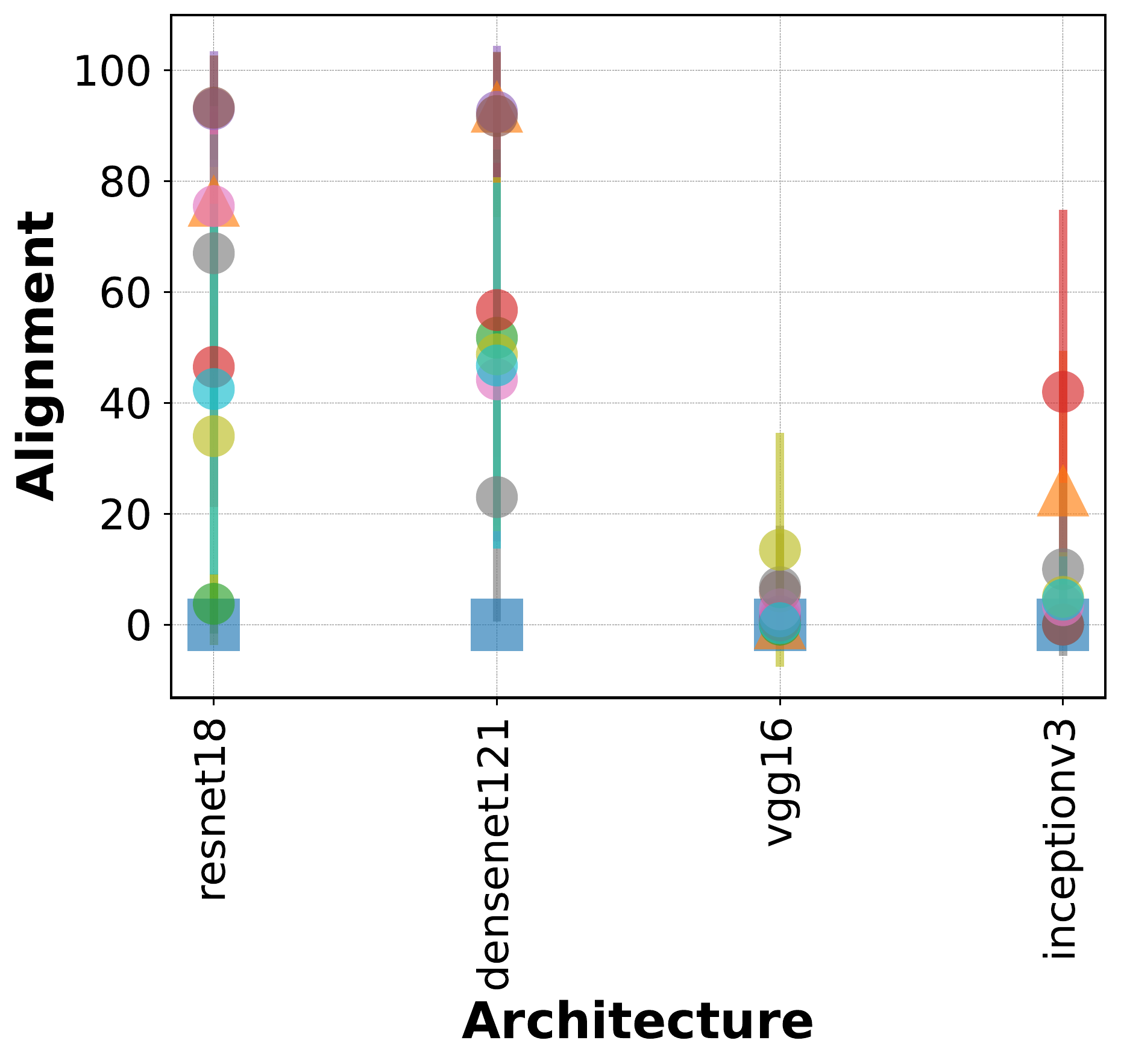}
        \end{subfigure}
        \raisebox{0.3\height}{\begin{subfigure}[b]{0.4\textwidth}
            \includegraphics[width=\textwidth]{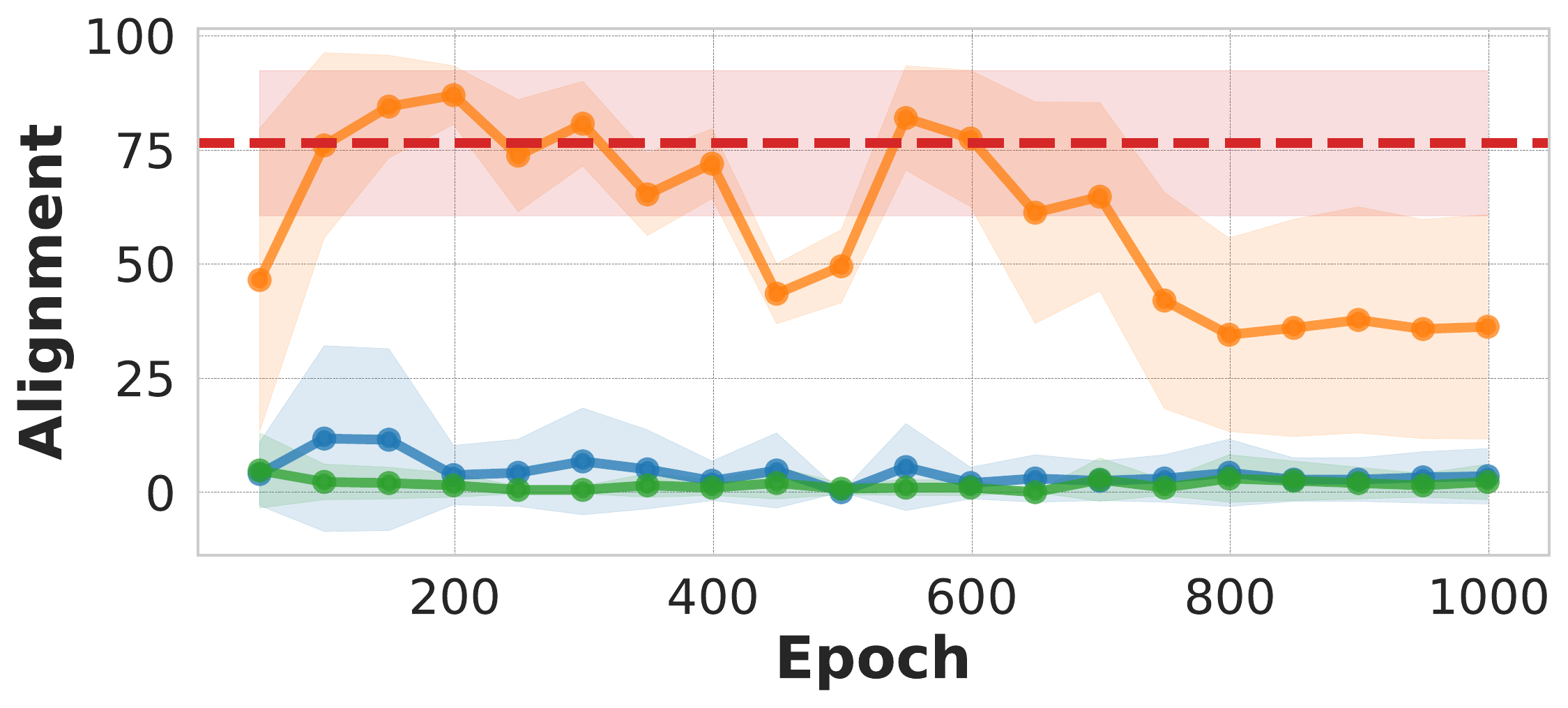}
        \end{subfigure}}
        
        \raisebox{0.6\height}{\begin{subfigure}[b]{0.4\textwidth}
            \includegraphics[width=\textwidth]{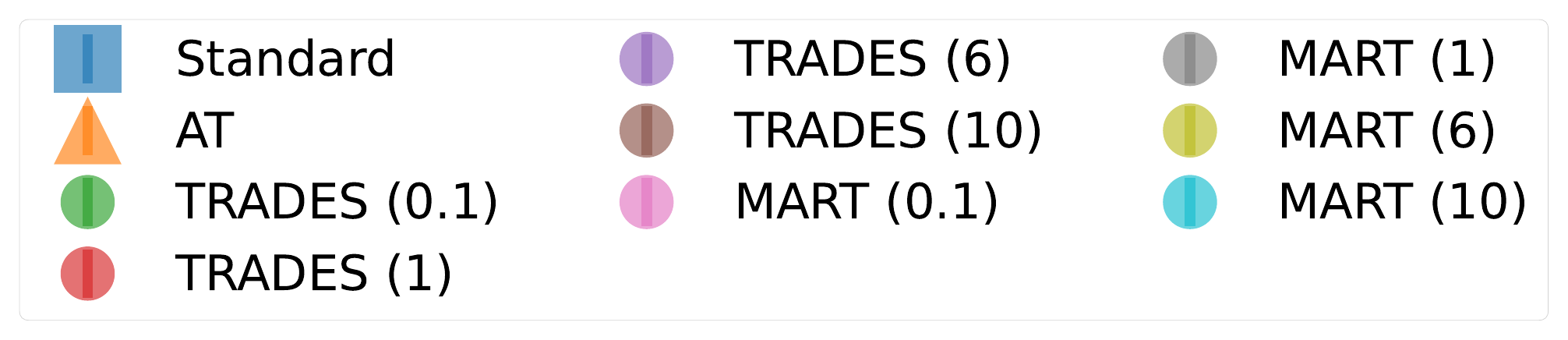}
            \caption{While adversarially robust models generally have high alignment, we see that different architectures. For example, VGG16 has very low levels of alignment despite being trained using robust training losses. Results on CIFAR100 and Imagenet in Appendix~\ref{sec:what_leads_appendix}.}
            \label{fig:loss_function_main}
        \end{subfigure}
        }
        \raisebox{0.7\height}{\begin{subfigure}[b]{0.4\textwidth}
            \includegraphics[width=\textwidth]{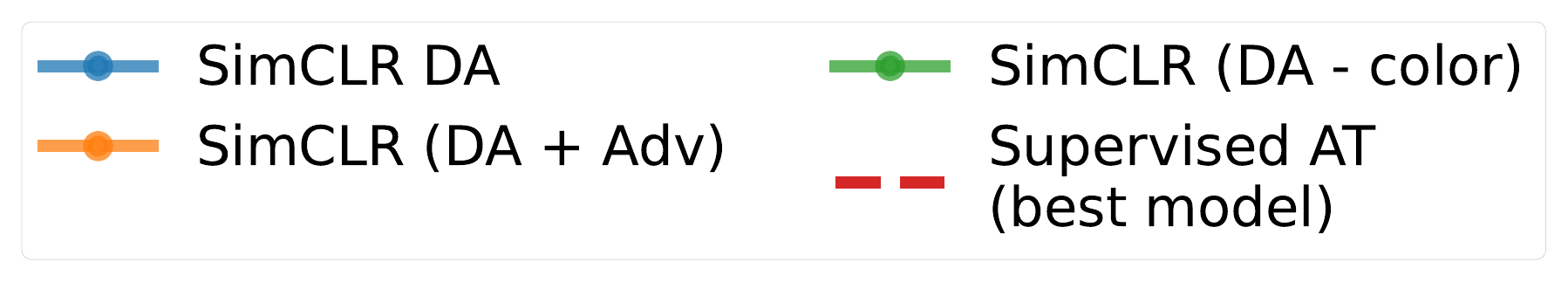}
            \caption{Combining SimCLR's data augmentations (DA) with adversarial augmentations (Adv) leads to best alignment (in the early and mid epochs) -- in some cases even surpassing the best supervised adversarially robust model. Results for more models and datasets in Appendix~\ref{sec:what_leads_appendix}.}
            \label{fig:learning_paradigm}
        \end{subfigure}
        }
    \vspace*{-20mm}
    \caption{Role of Loss Function in Alignment (left); Role of Training Paradigm in Alignment (right); ResNet18, CIFAR10}
    \vspace{-3mm}
    \label{fig:loss_function_and_simclr}
    \end{figure*}

\section{What Contributes to Learning Invariances Aligned with Humans}\label{sec:what_leads}

There have been many enhancements in the deep learning pipeline that have lead to remarkable generalization performance~\cite{krizhevsky2012imagenet,he2016deep,szegedy2015going,simonyan2014very,chen2020simple,kingma2014adam,ioffe2015batch}. In recent years there have been efforts to understand how invariances in representations learnt by such networks align with those of humans~\cite{geirhos2018imagenet,hermann2020origins,jenelle2019metamers}. However, how individual components of the deep learning pipeline affect the invariances learned is still not well understood. Prior works claim that adversarially robust models tend to learn representations with a ``human prior''~\cite{kaur2019perceptually, engstrom2019adversarial, santurkar2019image}. This leads to the question: how do other factors such as architecture, training paradigm, and data augmentation affect the invariances of representations? 

We explore these questions in this section. 
All evaluations in this section are based on regularizer-free \iris. We chose regularizer-free loss over the adversarial loss as the latter shows worst case alignment for all models, which is not useful for understanding the effect of various factors in the deep learning pipeline (Appendix~\ref{sec:regulziers_appendix} shows more results using the adversarial regularizer). 
Similarly, we preferred regularizer-free over human-leaning loss as the latter has a strong `bias' enforced by the regularizer.
While our approach generalizes to any layer, unless stated otherwise, all measurements of alignment are on the penultimate layer of the network.

\subsection{Architectures and Loss Functions}\label{sec:arch_and_loss_function}

We test the alignment of different DNNs trained using various loss functions -- standard cross-entropy loss, adversarial training (AT), and variants of AT (TRADES~\cite{zhang2019theoretically},  MART~\cite{wang2019improving}). Both TRADES and MART have two loss terms -- one each for clean and adversarial samples, which are balanced via a hyperparameter $\beta$. We report results for multiple values of $\beta$ in Fig~\ref{fig:loss_function_main} and find that the alignment of standard models (blue squares) is considerably worse than the robust ones (triangles and circles). However, the effect is also influenced by the choice of model architecture, \eg, for CIFAR10, for all  robust training losses, VGG16 has significantly lower alignment than other architectures.

\subsection{Data Augmentation}

Hand-crafted data augmentations are commonly used in deep learning pipelines. If adversarial training -- which augments adversarial samples during training -- generally leads to better aligned representations, then how do hand-crafted data augmentations affect invariances of learned representations? For adversarially trained models, we try with and without the usual data augmentation (horizontal flip, color jitter, and rotation). Since standard models trained with usual data augmentation show poor alignment (Section~\ref{sec:arch_and_loss_function}), we try stronger data augmentation (a composition of random flip, color jitter, grayscale and gaussian blur, as used in SimCLR~\cite{chen2020simple}) to see if hand-crafted data augmentations can improve alignment. Table~\ref{tab:data_aug} Appendix~\ref{sec:what_leads_appendix} shows how hand-crafted data augmentation can be crucial in learning aligned representations for some models (\eg, adversarially trained ResNet18 benefits greatly from data augmentation). In other cases data augmentation never hurts the alignment. We also see that standard models do not gain alignment even with stronger hand-crafted data augmentations. CIFAR100 and ImageNet results can be found in Table~\ref{tab:data_aug_appendix} Appendix~\ref{sec:what_leads_appendix} with similar takeaways.

\subsection{Learning Paradigm}

Since data augmentations (both adversarial and hand-crafted) along with residual architectures (like Resnet18) help alignment, self-supervised learning (SSL) models -- which explicitly rely on data augmentations -- should learn well aligned representations. This leads to a natural question: how do SSL models compare with the alignment of supervised models? 
SimCLR~\cite{chen2020simple} is a widely used contrastive SSL method that learns `meaningful' representations without using any labels. Recent works have built on SimCLR to also include adversarial data augmentations~\cite{kim2020adversarial,chen2020adversarial}. We train both the standard version of SimCLR (using a composition of transforms, as suggested in~\cite{chen2020adversarial}) and the one with adversarial augmentation on CIFAR10 and compare their alignment with the supervised counterparts. More training details are included in Appendix~\ref{sec:what_leads_appendix}. Additionally we also train SimCLR without the color distortion transforms -- which were identified as key transforms by the authors~\cite{chen2020simple} -- to see how transforms that are crucial for generalization affect alignment.
Fig~\ref{fig:learning_paradigm} shows the results when comparing self-supervised and supervised learning. We see that SimCLR when trained with both hand-crafted and adversarial augmentations has the best alignment, even outperforming the best adversarially trained supervised model in initial and middle epochs of training. We also see that removing color based augmentations (DA - color) does not have a significant impact on alignment, thus showing that certain DA can be crucial for generalization but not necessarily for alignment.

\xhdr{Summary} We find that there are three key components that lead to good alignment: architectures with residual connections, adversarial data augmentation using $\ell_2$ threat model, and a (self-supervised) contrastive loss. We leave a more comprehensive study of the effects of these training parameters on alignment for future work.

\section{Related Work}

\xhdr{Robust Models}
Several methods have been introduced to make deep learning models robust against adversarial attacks~\cite{papernot2016distillation,papernot2017practical,ross2018improving,gu2014towards,tramer2017ensemble,madry2019deep,zhang2019theoretically,wang2019improving,cohen2019certified}. These works try to model a certain type of human invariance (small change to input that does not change human perception) and make the model also learn such an invariance. Our work, on the other hand, aims to evaluate what invariances have already been learned by a model and how they align with human perception.
\xhdr{Representation Similarity}
There has been a long standing interest in comparing neural representations~\cite{laakso2000content,raghu2017svcca,morcos2018insights,kornblith2019similarity,wang2018towards,li2016convergent,nanda2022measuring,ding2021grounding}. 
While these works are related to ours in that they compare two systems of cognition, they assume complete white-box access to both neural networks. 
In our work, we wish to compare a DNN and a human, with only black-box access to the latter. 
\xhdr{DNNs and Human Perception} Neural networks have been used to model many perceptual properties such as quality~\cite{amirshahi2016image,gao2017deepsim} and  closeness~\cite{zhang2018perceptual} in the image space. Recently there has been interest in measuring the alignment of human and neural network perception. Roads et al. do this by eliciting similarity judgements from humans on ImageNet inputs and comparing it with the outputs of neural nets~\cite{roads21enriching}. Our work, however, explores alignment in the opposite direction, \ie, we measure if inputs that a network seed the same are also the same for humans. 
~\cite{jenelle2019metamers,engstrom2019adversarial} are closest to our work as they also evaluate alignment from model to humans, however as discussed in Section~\ref{sec:measuring_alignment}, unlike our work, their approaches are not scalable, they do not discuss the effects of loss function used to generate \iris, and they do not contribute to an understanding of what components in the deep learning pipeline lead to learning human-like invariances.

\section{Conclusion and Broader Impacts}\label{sec:broader_impact}

Our work offers insights into how measures of alignment can vary based on different loss functions used to generate \iris. We believe that when it is done carefully, measuring alignment is a useful model evaluation tool that provides insights beyond those offered by traditional metrics such as clean and robust accuracy, enabling better alignment of models with humans. We recognize that there are potentially worrying use cases against which we must be vigilant, such as taking advantage of alignment to advance work on deceiving humans. Human perception is complex, nuanced and discontinuous~\cite{stankiewicz1996categorical}, which poses many challenges in measuring the alignment of DNNs with human perception~\cite{guest2017success}. In this work, we take a step toward defining and measuring the alignment of DNNs with human perception. Our proposed method is a necessary but not sufficient condition for alignment and, thus, must be used carefully and supplemented with other checks, including domain expertise. By presenting this method, we hope for better design, understanding, and auditing of DNNs.

\section*{Acknowledgements}
AW acknowledges support from a Turing AI Fellowship under grant EP/V025279/1, The Alan Turing Institute, and the Leverhulme Trust via CFI. VN, AM, CK, and KPG were supported in part by an ERC Advanced Grant “Foundations for Fair Social Computing” (no. 789373). VN and JPD were supported in part by NSF CAREER Award IIS-1846237, NSF D-ISN Award \#2039862, NSF Award CCF-1852352, NIH R01 Award NLM013039-01, NIST MSE Award \#20126334, DARPA GARD \#HR00112020007, DoD WHS Award \#HQ003420F0035, ARPA-E Award \#4334192, ARL Award W911NF2120076 and a Google Faculty Research Award. BCL was supported by Wellcome Trust Investigator Award WT106931MA and Royal Society Wolfson Fellowship 183029. All authors would like to thank Nina Grgić-Hlača for help with setting up AMT surveys.

\bibliographystyle{plain}
\bibliography{main}

\clearpage
\appendix

\begin{table*}[t!]
\begin{center}
\begin{small}
\begin{sc}
\begin{tabular}{c|c|c}
\hline

\multicolumn{3}{c}{CIFAR10} \\
\hline

& \multirow{2}{*}{\begin{tabular}[c]{@{}c@{}}$x_0 \sim \mathcal{N}(0,1)$\end{tabular}} & \multirow{2}{*}{\begin{tabular}[c]{@{}c@{}}$x_0 \sim \mathcal{N}(0,0.01)$\end{tabular}} \\
& & \\
 \hline

\multirow{2}{*}{Model} & 
\multicolumn{2}{c}{\multirow{2}{*}{\begin{tabular}[c]{@{}c@{}}Alignment on Adversarial \iris \end{tabular}}} \\
 & \multicolumn{2}{c}{} \\
\hline

Supervised ResNet18, Standard & $1.33_{\pm 1.89}$ & $0.00_{\pm 0.00}$ \\
Supervised VGG16, Standard & $0.00_{\pm 0.00}$ & $0.00_{\pm 0.00}$ \\
Supervised InceptionV3, Standard & $0.33_{\pm 0.47}$ & $0.00_{\pm 0.00}$ \\
Supervised Densenet121, Standard & $0.00_{\pm 0.00}$ & $0.00_{\pm 0.00}$ \\
\hline
Supervised ResNet18, AT $\epsilon(\ell_2) = 1$ & $0.33_{\pm 0.47}$ & $1.00_{\pm 0.00}$ \\
Supervised VGG16, AT $\epsilon(\ell_2) = 1$ & $1.00_{\pm 1.41}$ & $1.00_{\pm 0.00}$ \\
Supervised InceptionV3, AT $\epsilon(\ell_2) = 1$ & $3.00_{\pm 4.24}$ & $0.00_{\pm 0.00}$ \\
Supervised Densenet121, AT $\epsilon(\ell_2) = 1$ & $1.33_{\pm 1.89}$ & $1.00_{\pm 0.00}$ \\
\hline
Supervised ResNet18, SimCLR DA & $0.00_{\pm 0.00}$ & $0.00_{\pm 0.00}$ \\
\hline
SimCLR ResNet18, Standard DA & $0.00_{\pm 0.00}$ & $1.00_{\pm 0.00}$ \\
SimCLR ResNet18, DA without Color & $0.00_{\pm 0.00}$ & $0.00_{\pm 0.00}$ \\
SimCLR ResNet18, Standard + Adv DA $\epsilon(\ell_2) = 1$ & $0.00_{\pm 0.00}$ & $0.00_{\pm 0.00}$ \\
 
\hline

\multicolumn{3}{c}{IMAGENET} \\
\hline

& \multirow{2}{*}{\begin{tabular}[c]{@{}c@{}}$x_0 \sim \mathcal{N}(0,1)$\end{tabular}} & \multirow{2}{*}{\begin{tabular}[c]{@{}c@{}}$x_0 \sim \mathcal{N}(0,0.01)$\end{tabular}} \\
& & \\
 \hline

\multirow{2}{*}{Model} & 
\multicolumn{2}{c}{\multirow{2}{*}{\begin{tabular}[c]{@{}c@{}}Alignment on Adversarial \iris \end{tabular}}} \\
 & \multicolumn{2}{c}{} \\
\hline

Supervised ResNet18, Standard & $0.00_{\pm 0.00}$ & $3.00_{\pm 0.00}$ \\
Supervised ResNet50, Standard & $0.00_{\pm 0.00}$ & $3.50_{\pm 0.50}$ \\
Supervised VGG16, Standard & $0.00_{\pm 0.00}$ & $3.00_{\pm 2.00}$ \\
\hline
Supervised ResNet18, AT $\epsilon(\ell_2) = 3$ & $0.33_{\pm 0.47}$ & $2.50_{\pm 1.50}$ \\
Supervised ResNet50, AT $\epsilon(\ell_2) = 3$ & $14.00_{\pm 3.74}$ & $3.50_{\pm 0.50}$ \\
Supervised VGG16, AT $\epsilon(\ell_2) = 3$ & $11.00_{\pm 3.74}$ & $4.50_{\pm 1.50}$ \\

\hline

\end{tabular}
\end{sc}
\end{small}
\end{center}

\caption{\textbf{[Adversarial \iris]} We observe that using the adversarial regularizer (described in Section~\ref{sec:generating_iris}) makes alignment for all models look bad. AT = Adversarial Training, DA = Data Augmentations. For details about standard SimCLR DA and DA without color, see Section~\ref{sec:what_leads_appendix}. }
\label{tab:adv_iris}
\end{table*}

\section{Measuring Human Alignment via Representation Inversion}\label{sec:human_alignment_appendix}

\subsection{Measuring Human Perception Similarity}
We recruited AMT workers with completion rate $\geq 95\%$ and who spoke English. To further ensure that the workers understood the task, we added attention checks. For 2AFC task, this meant making the query image as the same image as one of the images in the option. In clustering setting, this meant making the image on the row same as one of the images in the columns. All the workers who took our survey passed the attention checks.

We estimated a completion time of about $30$ minutes for each survey and thus paid each worker $7.5\$$. We allotted $60$ minutes per survey, so workers are not forced to rush through the survey. Most of the workers were able to complete the task in less than 20 minutes. Our study was approved by the Ethical Review Board of our institute.

\xhdr{ImageNet Clustering Hard} In order to create the hard ImageNet clustering task we use the human annotations of similarity between ImageNet images collected by ImageNet-HSJ authors~\cite{roads21enriching}. This contains a matrix ($M$) of similarity scores for each image in ImageNet validation set, where $M_{ij}$ is an indication for similarity between $i^{th}$ and $j^{th}$ images. For each image $i$ (randomly picked), we sample two more images that are the most similar to $i$ as per $M_i$. This creates a task that is much harder to perform for human annotators since the images on the columns look perceptually very similar (See Fig~\ref{fig:imagenet_clustering_hard} for an example).

\subsection{Using LPIPS as a proxy for $g_{\text{human}}$}\label{sec:g_human_appendix}
In order to ensure that LPIPS~\cite{zhang2018perceptual} is a reliable proxy to simulate human perception we measured if LPIPS could simulate human annotators on two perceptual similarity task setups: \texttt{2AFC} and \texttt{Clustering}, as described in Section~\ref{sec:g_human}. For \texttt{2AFC}, this meant using LPIPS to measure the distance between the query image and the two images shown in the options and then matching the query image to the one with lesser LPIPS distance. And similarly in \texttt{Clustering}, for each image on the row, we used LPIPS to measure its distance from each of the 3 images in the column and then matched it to the closest one. Our results (Table~\ref{tab:survey_results} in main paper) show that this can serve as a good proxy for human perception similarity. For all our experiments, we report avergae over 4 different LPIPS backbones: ImageNet trained Alexnet \& VGG16, and both of the Imagenet trained Alexnet \& VGG16 finetuned by the authors for perceptual similarity~\url{https://github.com/richzhang/PerceptualSimilarity}.

\begin{table*}[t!]
\begin{center}
\begin{small}
\begin{sc}
\begin{tabular}{c|c|c|c}
\hline

\multicolumn{4}{c}{CIFAR10} \\
\hline

& \multirow{2}{*}{\begin{tabular}[c]{@{}c@{}}In-\\Dist.\end{tabular}} & \multirow{2}{*}{\begin{tabular}[c]{@{}c@{}}Noise\\$\mathcal{N}(0,1)$\end{tabular}} & \multirow{2}{*}{\begin{tabular}[c]{@{}c@{}}Noise\\$\mathcal{N}(0.5,2)$\end{tabular}} \\
& & & \\
 \hline

\multirow{2}{*}{Model} & 
\multirow{2}{*}{\begin{tabular}[c]{@{}c@{}}Human\\2AFC\end{tabular}} &
\multirow{2}{*}{\begin{tabular}[c]{@{}c@{}}Human\\2AFC\end{tabular}} &
\multirow{2}{*}{\begin{tabular}[c]{@{}c@{}}Human\\2AFC\end{tabular}} \\
 & & & \\
\hline

ResNet18 & $96.00_{\pm 2.55}$ & $31.053_{\pm 15.912}$ & $64.386_{\pm 7.310}$ \\
VGG16 & $38.83_{\pm 7.59}$ & $0.351_{\pm 0.496}$ & $0.351_{\pm 0.496}$ \\
InceptionV3 & $82.00_{\pm 8.44}$ & $28.772_{\pm 20.476}$ & $2.105_{\pm 1.549}$ \\
Densenet121 & $98.67_{\pm 0.24}$ & $60.702_{\pm 11.413}$ & $68.947_{\pm 16.119}$ \\
 
\hline

\multicolumn{4}{c}{IMAGENET} \\
\hline

& \multirow{2}{*}{\begin{tabular}[c]{@{}c@{}}In-\\Dist.\end{tabular}} & \multirow{2}{*}{\begin{tabular}[c]{@{}c@{}}Noise\\$\mathcal{N}(0,1)$\end{tabular}} & \multirow{2}{*}{\begin{tabular}[c]{@{}c@{}}Noise\\$\mathcal{N}(0.5,2)$\end{tabular}} \\
& & & \\
 \hline

\multirow{2}{*}{Model} & 
\multirow{2}{*}{\begin{tabular}[c]{@{}c@{}}Human\\2AFC\end{tabular}} &
\multirow{2}{*}{\begin{tabular}[c]{@{}c@{}}Human\\2AFC\end{tabular}} &
\multirow{2}{*}{\begin{tabular}[c]{@{}c@{}}Human\\2AFC\end{tabular}} \\
 & & & \\
\hline

ResNet18 & $93.17_{\pm 5.95}$ & $28.748_{\pm 34.491}$ & $65.079_{\pm 5.616}$ \\
ResNet50 & $99.50_{\pm 0.00}$ & $70.745_{\pm 7.409}$ & $93.617_{\pm 9.027}$ \\
VGG16 & $95.50_{\pm 2.12}$ & $29.806_{\pm 19.704}$ & $46.914_{\pm 13.827}$ \\

\hline

\end{tabular}
\end{sc}
\end{small}
\end{center}

\caption{\textbf{[CIFAR10 and ImageNet In-Distr vs OOD Survey Results]} We observe that even on OOD samples that look like noise, humans can still bring out the relative differences between models, \eg, densenet121 on CIFAR10 is still ranked best aligned model on targets sampled from both kinds of noise. Reduced accuracy of humans on noise shows that this identifying similarities between \iris~on OOD samples is a harder task than with in-distribution target samples. }
\label{tab:in_vs_ood_survey}

\end{table*}

\subsection{Role of Input Distribution}\label{sec:choosing_x_appendix}

\begin{figure*}[t!]
    \centering
        \includegraphics[width=0.75\textwidth]{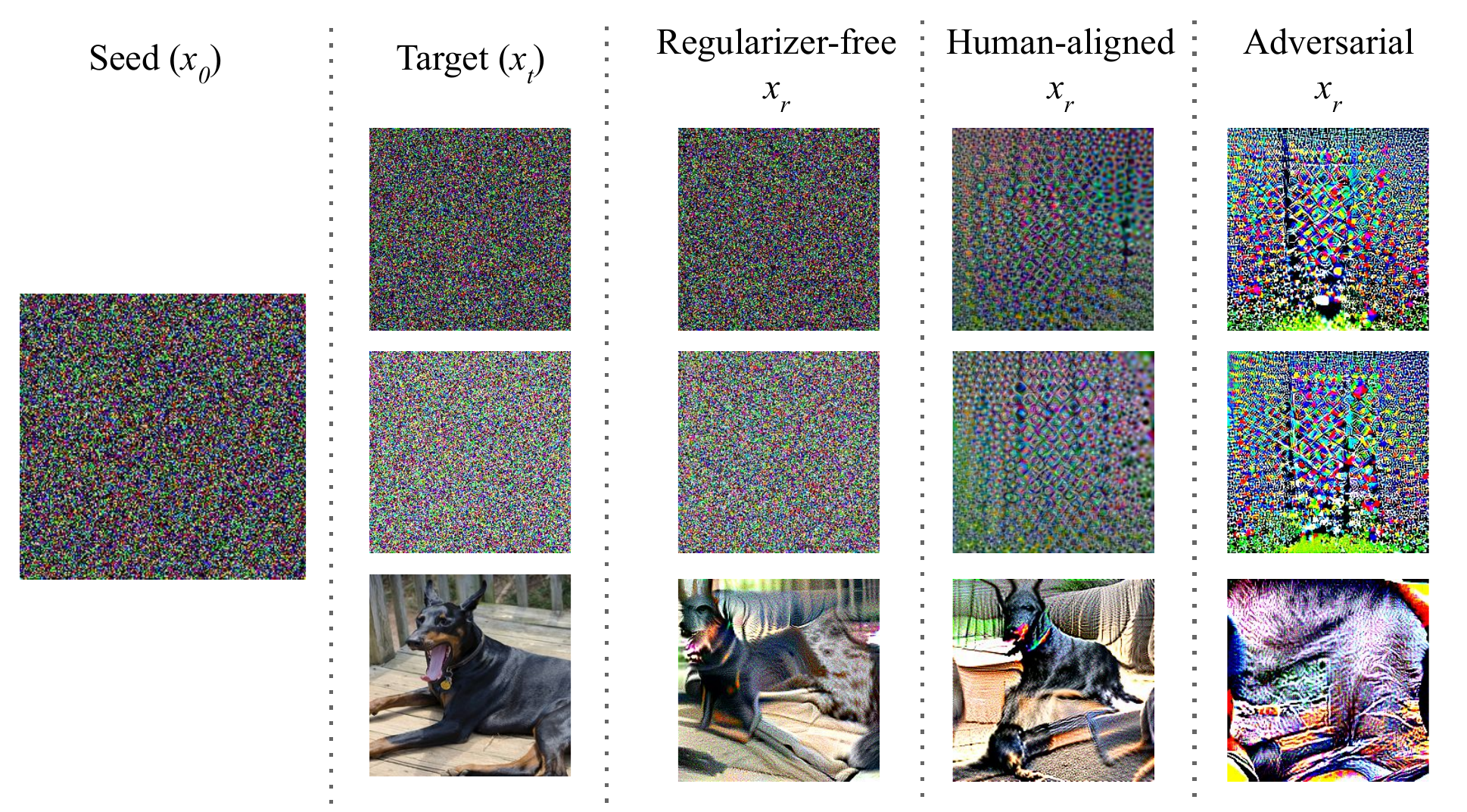}
        
\caption{ \textbf{[In vs Out of Distribution Samples]} Examples of reconstructions of in-distribution (bottom row) sample vs out-of-distribution samples for an ImageNet trained ResNet50 using the three regularizers mentioned in Section~\ref{sec:generating_iris}. Here the OOD targets are sampled from two separate random gaussians $\mathcal{N}(0, 1)$ (top row) and $\mathcal{N}(0.5, 2)$ (second row). We see that similar to the in-distribution sample, regularizer-free and adversarial inversions result in $x_r$ resembling and differing from $x_t$ respectively. Interestingly, for human-aligned regularizer, which explicitly tries to remove high-frequency features from $x_r$ fails to reconstruct an $x_t$ that itself consists of high-frequency features.
}
\label{fig:in_vs_ood_reconstructions}
\vspace{-6mm}
\end{figure*}

Fig~\ref{fig:in_vs_ood_reconstructions} shows some examples of inputs sampled from different gaussians (the lighter ones are sampled from $\mathcal{N}(0.5, 2)$ and darker ones from $\mathcal{N}(0, 1)$). We find that completing \texttt{2AFC} and \texttt{Clustering} tasks on inputs that look like noise to humans is a qualitatively harder task than when doing this on in-distribution target samples. 

However, remarkably, we observe that humans are still able to bring out the differences between different models, even when given (a harder) task of matching re-constructed noisy inputs. Table~\ref{tab:in_vs_ood_survey} shows the results of surveys conducted with noisy target samples. It's worth noting that the accuracy of humans drop quite a bit from in-distribution targets, thus indicating that this is indeed a harder task.

\subsection{Regularizers}\label{sec:regulziers_appendix}

For the human-aligned regularizers, we use the ones discussed in~\cite{olah2017feature}. These fall into three broad categories: \textit{frequency penalization}, \textit{transformation robustness}, and \textit{pre-conditioning}.
\begin{itemize}
    \item \textbf{Frequency Penalization}: The goal is to explicitly penalize high-frequency features in the reconstruction ($x_r$). This is done by adding a regularizer of the form $\mathcal{R}(x) = TV(x) + ||x||_p$, where $TV$ is the total variation and p = 1~\cite{mahendran2014understanding}. A similar effect of frequency penalization can also be done by ensuring robustness of $x_r$ to blurring, \ie, $\mathcal{R}(x) = || x - \text{StopGradient}(blur(x)) ||_2^{2}$~\cite{Nguyen2015deep}.
    \item \textbf{Transformation Robustness}: This ensures that $x_r$ is such that the representation is same even if we slightly transform $x_r$. This is achieved by replacing $x$ with $T(x)$ in Eq~\ref{eq:inversion}. We use T as a composition of color jitter, random scaling, and random rotation.
    \item \textbf{Pre-conditioning}: This involves taking gradient steps in the fourier domain, which decorrelates the pixels in $x_r$. 
\end{itemize}

For our experiments we find that transformation robustness generates the best looking $x_r$ and thus we report results under \textit{human-learning regularizer} based on $x_r$ generated using \textit{transformation robustness} during representation inversion.

For adversarial regularizer, we report results in Table~\ref{tab:adv_iris} and find that such a regularizer can make almost all models look like they have bad alignment.

\subsection{Role of $x_0$}\label{sec:role_of_seed_appendix}

We additionally report results for the adversarial regularizer where \iris~were generated from a separate seed. While experiments in the main paper reported for a seed sampled from $\mathcal{N}(0, 1)$, we report results here for a seed sampled from $\mathcal{N}(0, 0.01)$ in Table~\ref{tab:adv_iris} and find that regardless of seed, adversarial regularizer makes all models look bad.

\section{Model, Code, Assets, and Compute Details}\label{sec:model_code_data_etc_appendix}

\subsection{Code and Assets}
In our code we make use of many open source libraries such as timm~\cite{rw2019timm}, pytorch~\cite{paszke2019pytorch}, pytorch-lightning~\cite{falcon2019pytl}, numpy~\cite{harris2020array}, robustness~\cite{robustness}, matplotlib~\cite{Hunter2007matplotlib}. timm, pytorch-lightning and have an Apache 2.0 license. Numpy has a BSD 3-Clause License. Robustness has an MIT license. PyTorch's license can be found here: \url{https://github.com/pytorch/pytorch/blob/master/LICENSE}, and matplotlib's here: \url{https://github.com/matplotlib/matplotlib/blob/main/LICENSE/LICENSE}. All these licenses allow free use, modification and distribution.
We use publicly available academic datasets CIFAR10/100~\cite{krizhevsky2009learning} and ImageNet~\cite{russakovsky2015imagenet}.

\subsection{Models}\label{sec:models_appendix}

\xhdr{Supervised} We used VGG16, ResNet18, Densenet121 and InceptionV3 for experiments on CIFAR10 and CIFAR100. The ``robust'' version of these models were trained using adversarial training~\cite{madry2019deep}, with an $\ell_2, \epsilon$ of $1$. All these models were trained using the standard data augmentations(a composition of \texttt{RandomCrop}, \texttt{RandomHorizontalFlip}, \texttt{ColorJitter}, \texttt{RandomRotation}). For ImageNet, we used VGG16, ResNet18 and ResNet50 and the ``robust'' versions of these models were taken from~\cite{salman2020adversarially} with an $\ell_2, \epsilon$ of $3$. ImageNet models used sightly different data augmentationsRandomHorizontalFlip
ColorJitter
Lighting

\xhdr{Self-supervised} We used SimCLR~\cite{chen2020simple} to train a ResNet18 backbone on CIFAR10 and CIFAR100. More details about different types of data augmentations in Section~\ref{sec:what_leads_appendix}. 

\xhdr{ImageNet} We used VGG16 (with batchnorm), ResNet18 and ResNet50 for ImageNet. The ``robust'' versions of these models were taken from~\cite{salman2020adversarially}, who trained these models using adversarial training with an $\ell_2$ epsilon of 3.

\subsection{Compute Details}
We used our institute's GPU cluster to run all experiments. Since our experiments involve standard models and datasets, these can be run on any hardware supported by PyTorch. In our case, we used 5 machines with 2 V100 Nvidia Tesla GPUs (32GB each, volta architecture) and a Nvidia dgx machine with 8 Nvidia Tesla P100 GPUs (16GB each). We estimate a total of 500+ GPU hours.

\section{What Contributes to Good Alignment}\label{sec:what_leads_appendix}

\begin{table*}[t!]
\captionsetup{font=footnotesize}
\begin{center}
\begin{small}
\begin{sc}
\begin{tabular}{c|cccc}
\hline

\multicolumn{5}{c}{Adv. Training} \\
\hline

& Resnet18 & Densenet121 & VGG16 & InceptionV3  \\
 \hline

\multirow{2}{*}{Usual Data Aug} & \multirow{2}{*}{$76.50_{\pm 15.91}$} & \multirow{2}{*}{$93.50_{\pm 9.60}$}  & \multirow{2}{*}{$0.25_{\pm 0.43}$} & \multirow{2}{*}{$24.25_{\pm 25.17}$} \\
 &  &  &  & \\
\multirow{2}{*}{No Data Aug} & \multirow{2}{*}{$30.00_{\pm 12.02}$} & \multirow{2}{*}{$93.75_{\pm 8.20}$}  & \multirow{2}{*}{$1.00_{\pm 1.73}$} & \multirow{2}{*}{$12.25_{\pm 20.08}$} \\
 &  &  &  & \\
\hline

\multicolumn{5}{c}{Standard} \\
\hline

& Resnet18 & Densenet121 & VGG16 & InceptionV3  \\
 \hline

\multirow{2}{*}{Strong Data Aug} & \multirow{2}{*}{$0.00_{\pm 0.00}$} & \multirow{2}{*}{$1.00_{\pm 1.73}$}  & \multirow{2}{*}{$0.00_{\pm 0.00}$} & \multirow{2}{*}{$0.00_{\pm 0.00}$} \\
 &  &  &  & \\
\multirow{2}{*}{Usual Data Aug} & \multirow{2}{*}{$0.00_{\pm 0.00}$} & \multirow{2}{*}{$0.00_{\pm 0.00}$} & \multirow{2}{*}{$0.00_{\pm 0.00}$} & \multirow{2}{*}{$0.00_{\pm 0.00}$} \\
 &  &  &  & \\
\hline

\end{tabular}
\end{sc}
\end{small}
\end{center}

\caption{\textbf{[CIFAR10 Models; Effect of Data Augmentation]} For certain models, \eg, adversarially trained resnet18, data augmentation is crucial in learning aligned representations.}
\label{tab:data_aug}

\end{table*}

\xhdr{SimCLR training details} We used data augmentations shown to work best by the authors (a composition of \texttt{RandomHorizontalFlip}, \texttt{ColorJitter}, \texttt{RandomGrayscale} and \texttt{GaussianBlur}, as implemented in the original codebase~\url{https://github.com/google-research/simclr}). We also train SimCLR models without the color augmentations (\ie~only \texttt{RandomHorizontalFlip} and \texttt{GaussianBlur}). Since color transforms were crucial for obtaining representations with good generalization performance, we wanted to analyze how removing augmentations crucial for generalization impacts alignment. Finally, we also train a variant of SimCLR with adversarial data augmentations, as proposed in some recent works~\cite{kim2020adversarial,chen2020adversarial}. As opposed to traditional adversarial training, here we generate adversarial data augmentations for a model ($g$) by solving the following maximization for each input $x$:
\begin{equation*}
    \text{argmax}_{x'} || g(x') - g(x) ||_2 \,\,\, \text{st} ||x - x'||_2 \leq \epsilon
\end{equation*}

For our experiments $\epsilon = 1$ for CIFAR10 and $\epsilon = 3$ for ImageNet (similar to supervised models).

\begin{figure*}
        \centering
        \begin{subfigure}[b]{0.3\textwidth}
            \includegraphics[width=\textwidth]{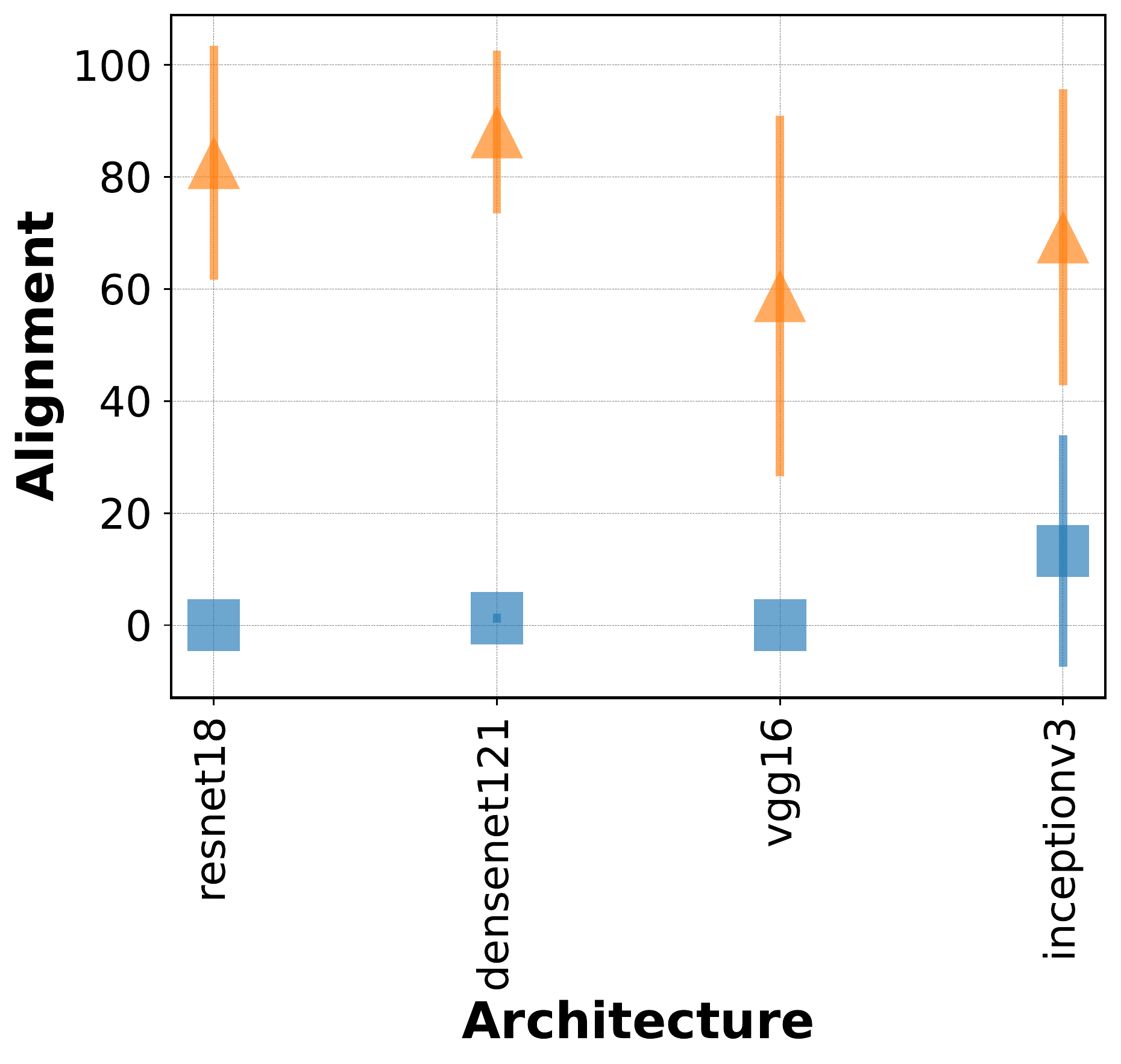}
        \end{subfigure}
        \hfill
        \raisebox{0.05\height}{\begin{subfigure}[b]{0.3\textwidth}
            \includegraphics[width=\textwidth]{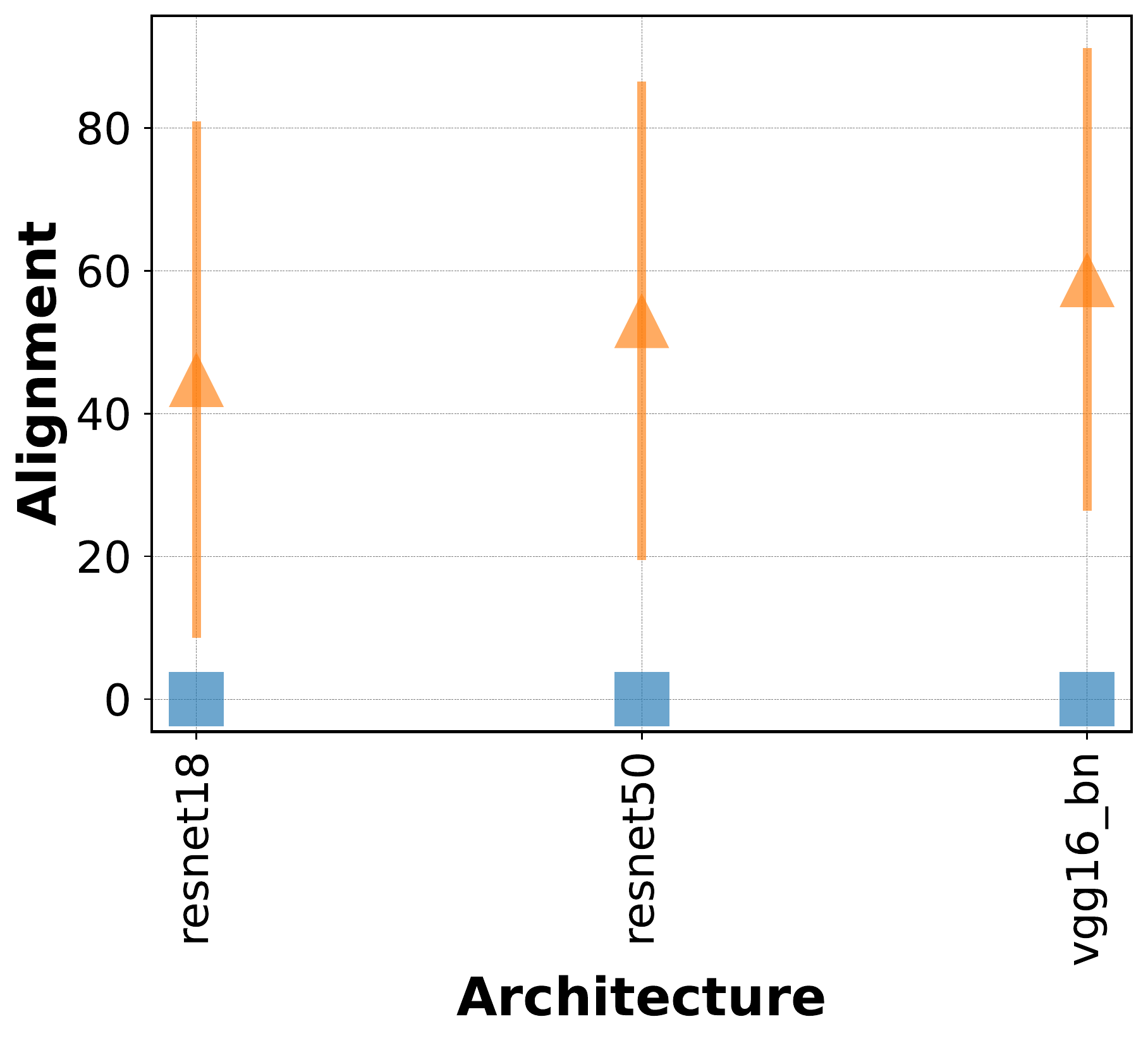}
        \end{subfigure}}
        
        \begin{subfigure}[b]{0.25\textwidth}
            \includegraphics[width=\textwidth]{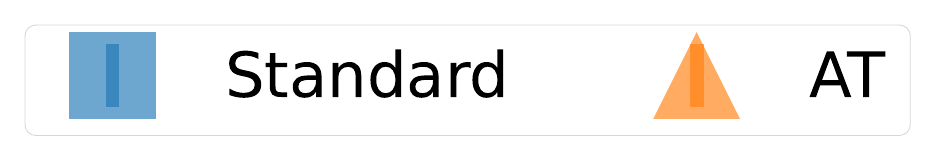}
        \end{subfigure}
    
    \caption{Role of Loss Function in Alignment; CIFAR100 (left), and ImageNet (right)}
    \label{fig:loss_function_appendix}
\end{figure*}

\xhdr{Architectures and Loss Function, CIFAR100 \& ImageNet} Fig~\ref{fig:loss_function_appendix} shows results for CIFAR100 and ImageNet for standard and robust training. Similar to previous works, we find that robust models are better aligned with human perception. Interestingly, we find that the variance between different architectures that we observed for CIFAR10 does not exist for CIFAR100 and ImageNet, \ie, regardless of architecture, robustly trained models are well aligned with human perception. Indicating that (unsurprisingly) training dataset also plays a major role in alignment.

\begin{figure*}
    \centering
    \begin{subfigure}[b]{0.4\textwidth}
        \includegraphics[width=\textwidth]{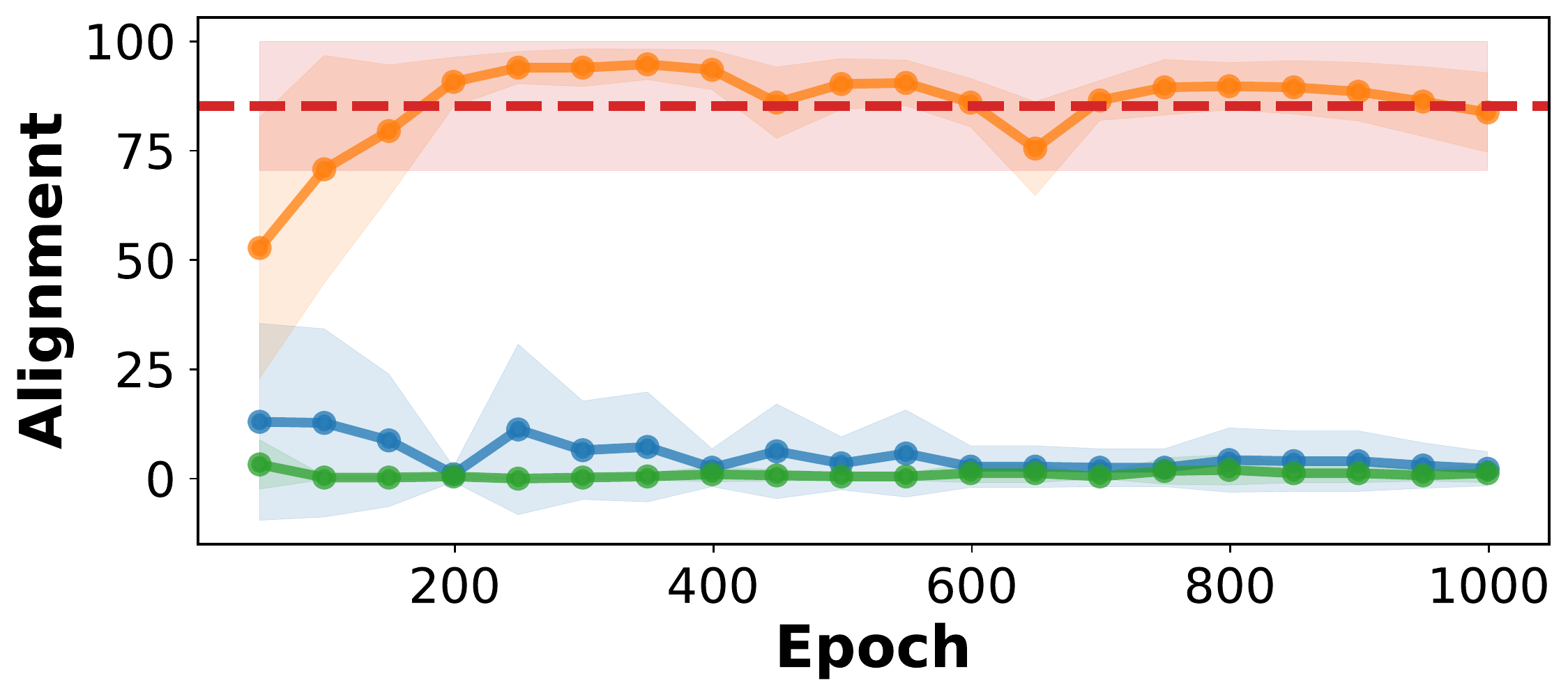}
    \end{subfigure}
    \raisebox{0.45\height}{\begin{subfigure}[b]{0.5\textwidth}
        \includegraphics[width=\textwidth]{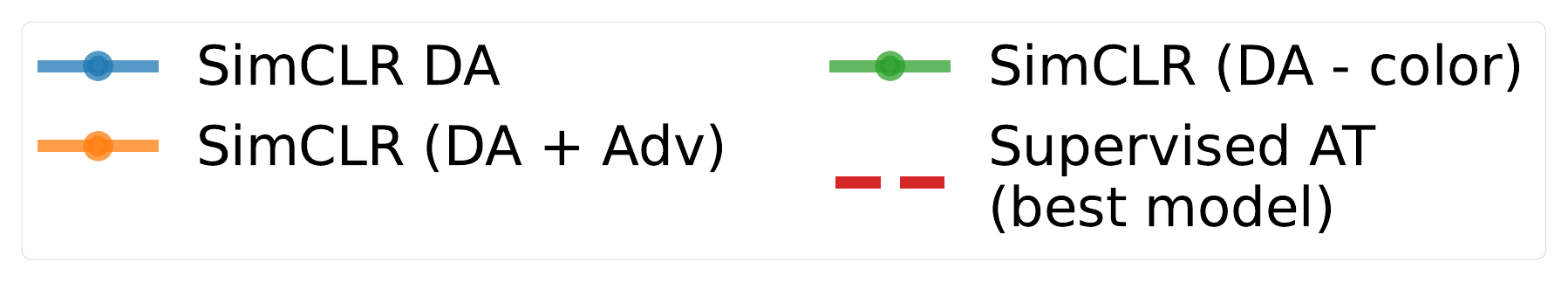}
    \end{subfigure}}
\caption{ResNet18 backbone trained using SimCLR on CIFAR100.}
\label{fig:simclr_appendix}
\end{figure*}

\xhdr{SimCLR, CIFAR100} Fig~\ref{fig:simclr_appendix} shows alignment of different types of SimCLR models throughout training. We observe a similar trend as CIFAR10, where adversarial data augmentation improves alignment.

\subsection{Data Augmentation, CIFAR10, CIFAR100 \& ImageNet}
Since re-training ImageNet models with adversarial training is very resource intensive, we train ImageNet models using Free Adversarial Training (Free AT)~\cite{shafahi2019adversarial}. Free AT only has an implementation for $\ell_{\text{inf}}$ threat model, hence we train these models with $\ell_{\text{inf}}$, $\epsilon=4/255$ (for both with and without data augmentation). Table~\ref{tab:data_aug_appendix} shows that similar to CIFAR10 (Table~\ref{tab:data_aug}), for some models, like ResNet18, data augmentation is crucial in learning aligned representations (despite being trained to be adversarially robust). For other models, data augmentation never hurts alignment (except InceptionV3 for CIFAR100).

\begin{table*}[b!]
\vspace{-3mm}
\captionsetup{font=footnotesize}
\begin{center}
\begin{small}
\begin{sc}
\begin{tabular}{c|cccc}
\hline

\multicolumn{5}{c}{CIFAR100} \\
\hline

& Resnet18 & Densenet121 & VGG16 & InceptionV3  \\
 \hline

\multirow{2}{*}{Usual Data Aug} & \multirow{2}{*}{$82.50_{\pm 20.85}$} & \multirow{2}{*}{$88.00_{\pm 14.51}$}  & \multirow{2}{*}{$58.75_{\pm 32.15}$} & \multirow{2}{*}{$69.25_{\pm 26.39}$} \\
 &  &  &  & \\
\multirow{2}{*}{No Data Aug} & \multirow{2}{*}{$81.50_{\pm 15.58}$} & \multirow{2}{*}{$89.75_{\pm 15.01}$} & \multirow{2}{*}{$46.25_{\pm 32.25}$} & \multirow{2}{*}{$84.75_{\pm 13.70}$} \\
 &  &  &  & \\
\hline

\multicolumn{5}{c}{ImageNet} \\
\hline

& \multicolumn{2}{c}{Resnet18} & \multicolumn{2}{c}{Resnet50} \\
 \hline

\multirow{2}{*}{Usual Data Aug} & \multicolumn{2}{c}{\multirow{2}{*}{$13.00_{\pm 22.52}$}} & \multicolumn{2}{c}{\multirow{2}{*}{$0.00_{\pm 0.00}$}} \\
 &  &  &  & \\
\multirow{2}{*}{No Data Aug} & \multicolumn{2}{c}{\multirow{2}{*}{$0.75_{\pm 1.30}$}} & \multicolumn{2}{c}{\multirow{2}{*}{$0.00_{\pm 0.00}$}} \\
 &  &  &  & \\
\hline

\end{tabular}
\end{sc}
\end{small}
\end{center}

\caption{\textbf{[CIFAR100 \& ImageNet Models all trained to be adversarially robust; Effect of Data Augmentation]} Similar to CIFAR10, data augmentation is crucial for adversarially trained resnet18 to learn aligned representations.}
\label{tab:data_aug_appendix}

\end{table*}

\end{document}